%% file: neurips_2025.tex
\documentclass{article}

\usepackage[preprint]{neurips_2025}

\usepackage[utf8]{inputenc} 
\usepackage[T1]{fontenc}    
\usepackage{hyperref}       
\usepackage{url}            
\usepackage{booktabs}       
\usepackage{amsfonts}       
\usepackage{nicefrac}       
\usepackage{microtype}      
\usepackage{xcolor}         

\usepackage{amsmath}
\usepackage{amssymb}
\usepackage{mathtools}
\usepackage{amsthm}

\definecolor{MyLightBlue}{RGB}{145, 239, 239}
\definecolor{MyDarkBlue}{RGB}{47, 85, 151}
\definecolor{MyDarkGreen}{rgb}{0.02,0.6,0.02}
\definecolor{MyDarkRed}{rgb}{0.8,0.02,0.02}
\definecolor{MyDarkOrange}{rgb}{0.40,0.2,0.02}
\definecolor{MyLightYellow}{RGB}{198, 204, 143}
\definecolor{MyRed}{rgb}{1.0,0.0,0.0}
\definecolor{MyGold}{rgb}{0.75,0.6,0.12}
\definecolor{MyDarkgray}{rgb}{0.66, 0.66, 0.66}

\usepackage{wrapfig}
\usepackage{subcaption}
\usepackage{multirow}
\usepackage{multicol}
\usepackage{enumitem}

\definecolor{mygray}{gray}{0.4}

\usepackage{longtable}
\usepackage{listings}
\usepackage{algorithm}
\usepackage{algorithmic}
\usepackage{hyperref}
\usepackage{url}
\usepackage{caption}
\title{Ella: Embodied Social Agents with Lifelong Memory}

\author{
Hongxin Zhang$^{1}$\thanks{denotes equal contribution.},
Zheyuan Zhang$^{2*}$,
Zeyuan Wang$^{3*}$,\\
\textbf{Zunzhe Zhang$^3$,}
\textbf{Qinhong Zhou$^{1}$,}
\textbf{Lixing Fang$^{1}$,}
\textbf{Chuang Gan$^{1}$} \\
$^1$ University of Massachusetts Amherst $^2$ Johns Hopkins University $^3$ Tsinghua University}

\begin{document}

\maketitle

\begin{abstract}
We introduce \textit{Ella}, an embodied social agent capable of lifelong learning within a community in a 3D open world, where agents accumulate experiences and acquire knowledge through everyday visual observations and social interactions.
At the core of \textit{Ella}'s capabilities is a structured long-term multimodal memory system that stores, updates, and retrieves information effectively. It consists of a name-centric semantic memory for organizing acquired knowledge and a spatiotemporal episodic memory for capturing multimodal experiences. 
By integrating this lifelong memory system with foundation models, \textit{Ella} retrieves relevant information for decision-making, plans daily activities, builds social relationships, and evolves autonomously while coexisting with other intelligent beings in the open world.
We conduct capability-oriented evaluations in a dynamic 3D open world where 15 agents engage in social activities for days and are assessed with a suite of unseen controlled evaluations. Experimental results show that \textit{Ella} can influence, lead, and cooperate with other agents well to achieve goals, showcasing its ability to learn effectively through observation and social interaction. 
Our findings highlight the transformative potential of combining structured memory systems with foundation models for advancing embodied intelligence. More videos can be found at \url{https://umass-embodied-agi.github.io/Ella/}.

\end{abstract}


\section{Introduction}

It's a long-standing goal to create intelligent beings capable of survival in the human community~\citep{gan2021threedworld, li2023behavior, puig2024habitat}, which requires lifelong learning in an open and social world. The embodied agents must accumulate experiences, including visual observations and social interactions with other intelligent beings, such as conversations; and acquire knowledge from these multi-modal experiences, build new concepts of objects, agents, and events, and identify the connections among these concepts.

\begin{figure}[t]
    \centering
    \begin{subfigure}[b]{0.5\linewidth}
        \centering
        \raisebox{0.5em}{ 
            \includegraphics[height=5.0cm]{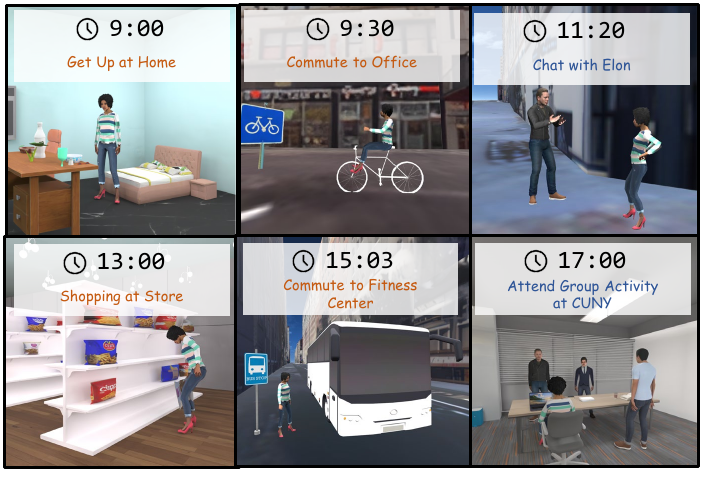}
        }
        \caption{}
        \label{fig:teaser}
    \end{subfigure}
    \hfill
    \begin{subfigure}[b]{0.45\linewidth}
        \centering
        \includegraphics[height=4.5cm]{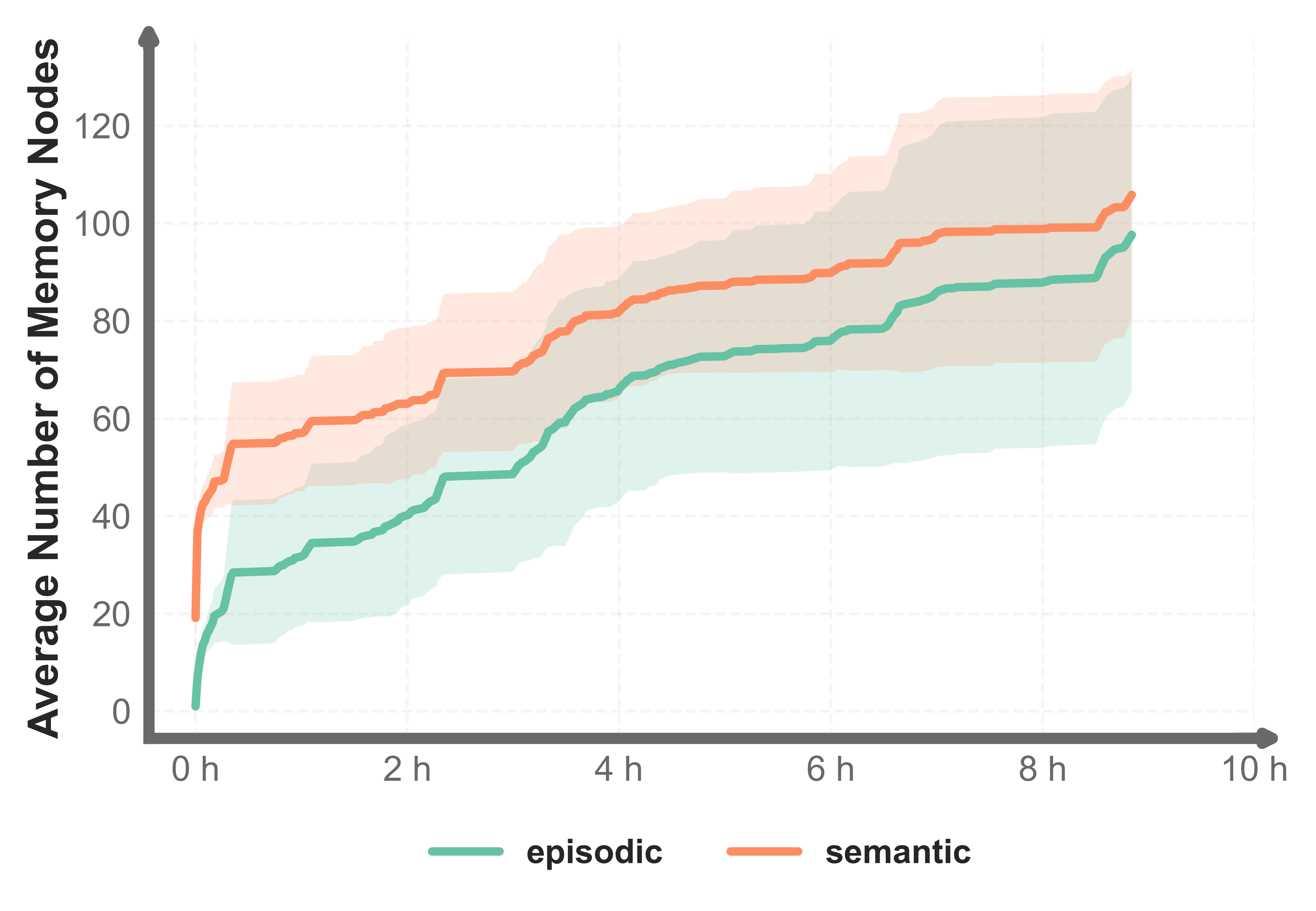}
        \caption{}
        \label{fig:memory_growth}
    \end{subfigure}
    \caption{(a) Embodied agents require lifelong learning to accumulate experiences and acquire knowledge through everyday visual observation and social interaction within a community in a 3D open world. (b) \textit{Ella} self-evolves by growing episodic and semantic memory over time.}
    \label{fig:combined}
    \vspace{-5mm}
\end{figure}

With the rapid advancement of Foundation Models~\citep{openai2023gpt4, ravi2024sam2segmentimages, guo2025deepseek}, a surge of powerful agents has emerged~\citep{sumers2023cognitive}. These range from agents operating solely in the text domain~\citep{gur2023real, shinn2024reflexion} to multi-modal agents capable of controlling screens~\citep{hong2024cogagent}, playing games~\citep{wang2023voyager, wang2023describe}, and even functioning as robots in the physical world~\citep{ahn2022can, huang2023voxposer, du2023video}. Despite these advancements, one crucial component remains underexplored in current agent research: long-term memory. Humans organize accumulated experiences in Episodic Memory~\citep{tulving1972episodic, tulving1983elements,nuxoll2007extending} and acquired knowledge in Semantic Memory\citep{lindes2016toward}, enabling them to make long-term plans and exhibit higher-level cognitive capabilities~\citep{laird2022introduction, tenenbaum2011grow}. In contrast, current work in embodied agents is limited to constrained spatial regions (primarily indoor spaces) and brief temporal scales (seconds for robotic manipulation or minutes for navigation tasks). For agents to thrive in an ever-evolving world, it is essential to develop a long-term memory system that supports learning new concepts and forming new relationships. In this direction, Generative Agents~\citep{park2023generative} introduced a textual temporal episodic memory, assuming oracle perception in a sandbox 2D environment. Similarly, Voyager~\citep{wang2023voyager} designed a single agent with long-term procedural memory, enabling it to acquire new skills in Minecraft through oracle perception and self-training. However, the challenge of constructing effective lifelong memory systems for embodied agents in an open and social world—where they must learn from visual observations and engage in social interactions with other intelligent beings, as illustrated in Figure~\ref{fig:teaser}—remains largely unexplored.

In this work, we propose to build a robust long-term multi-modal memory system that can store, update, and retrieve information effectively. Borrowing the concepts from psychology and cognitive neuroscience~\citep{tulving1972episodic}, we construct the long-term memory in two forms: a name-centric semantic memory with a hierarchical scene graph and knowledge graph to organize acquired knowledge, and a spatiotemporal episodic memory to capture the agent's multi-modal experiences. We present \textit{Ella}, an embodied lifelong learning agent that can accumulate experiences and acquire knowledge effectively through visual perception and social interaction with other agents within a community in an open 3D world, by integrating this structured memory with foundation models. To plan robustly and behave consistently through days of social life, \textit{Ella} adopts a planning-reaction framework where it first retrieves related context from memory to make a structured daily schedule, then updates the memory with new visual observations and social interactions, and make reactions to the new context, which could be revising the schedule, interacting with the environment, or engaging in social interactions. 

We simulate \textit{Ella} and other baseline agents in \textit{Virtual Community}~\citep{vico}, an open world simulation platform for multi-agent embodied AI, featuring large-scale community scenarios derived from the real world with realistic physics and renderings. Unlike traditional task-oriented evaluations for agents, assessing high-level cognitive capabilities in a lifelong setting is more critical~\citep{crosby2019animal}. To this end, we first simulate 15 agents for 9 hours (with a decision frequency of 1 second), representing their first day in the community. During this phase, agents must plan their day based on their unique characteristics and acclimate to the environment and other agents. Then we test the agents with unseen controlled evaluations: \textit{Influence Battle} and \textit{Leadership Quest}, where the agents work in groups to persuade others to attend their party at a specific location despite conflicting schedules or lead their group to prepare for an activity under resource constraints. Experimental results across three communities show that \textit{Ella} demonstrates advanced cognitive abilities including social reasoning and leadership, showcasing its ability to learn effectively through visual observation and social interaction. In sum, our contribution includes:

\begin{itemize}
    \item We propose a structured long-term memory with name-centric semantic memory and spatiotemporal episodic memory to support lifelong learning in an open and social world.

   \item We introduce \textbf{\textit{Ella}}, the first embodied social agent that can self-evolve through visual observation and social interaction by integrating structured memory with foundation models.
    
    \item We conduct capability-oriented experiments in a dynamic 3D open world with 15 agents for days, and demonstrate \textbf{\textit{Ella}}'s advanced cognitive abilities including social reasoning and leadership.

\end{itemize}

\section{Related Work}

\subsection{Embodied Social Intelligence}
Social intelligence has been widely studied in embodied multi-agent environments~\citep{vico,lowe2017multi,carroll2019utility, amato2019modeling,bard2020hanabi, jain2020cordial, puigwatch, tsoi2020sean, puig2023habitat, wen2022multi, szot2023adaptive, zhang2023building, li2019robust}, while one branch focuses on simplified symbolic or game-like environments~\citep{samvelyan2019starcraft,suarez2019neural,jaderberg2019human,Baker2020Emergent, niu2021multi, sharon2015conflict, yu2024mineland}, often ignoring the challenges present in an open world, including perception and diverse personalities of agents.
Specifically, generative agents~\citep{park2023generative} developed a unified temporal language memory, demonstrating the robust simulation of human-like agents within a symbolic community. Following this line of research, a series of works have explored socially intelligent agents within text-based sandbox environments~\cite{li2023camel, zhou2024sotopia, liu2024training, chen2024socialbench, liu2024exploring, liu2024interintent, dai2024artificial}.
The other branch, including works on human-robot interaction~\citep{gombolay2015decision, goodrich2008human,bobu2023aligning,dautenhahn2007socially,nikolaidis2015efficient,rozo2016learning,losey2022learning, natarajan2020effects, lasota2017survey}, focuses on real-world domains but is limited to specific task settings.
Different from above, we explore embodied social intelligence within a community in an open 3D world, featuring expansive spatial regions and a temporal scale spanning multiple days.

\subsection{Agent Memory}
Memory has been studied for 
a long time in AI, especially related to cognitive architectures~\citep{weston2014memory,lindes2016toward,sumers2023cognitive}. However, most modern agent architecture primarily assumes a temporal memory due to the constraints of specific domains or the limited time horizon for which the agent is designed.
A visual memory as a type of semantic memory has been implemented using various structures in computer vision, including voxels~\citep{chaplot2020object, blukis2022persistent, min2022film, ramakrishnan2022poni}, scene graphs~\citep{li2022embodied, rana2023sayplan, kurenkov2023modeling, gu2024conceptgraphs}, Octrees~\citep{hornung2013octomap, zhang2018semantic, asgharivaskasi2023semantic, zheng2023asystem}, or implicit continuous representations~\citep{shafiullah2022clip, gadre2022continuous, huang2023visual, gadre2023cows}. 
Recently, several works have explored agent memory for longer time horizons. \citep{kurenkov2023modeling, yang2024snapmem} and \citep{zhou2023long} introduce updating mechanisms for scene graph-based memory, adapting it to long-term tasks. \citep{wang2023voyager} and \citep{li2024optimus} develop procedural memory tailored for specific game environments to support long-term planning. \citep{jiang2024long} proposes long-term memory with a graph-based structure to enable self-evolution in LLM tasks. \citep{wang2024karma} further integrates long-term and short-term memory to address long-horizon tasks within household environments. Another line of work studies how to better retrieve knowledge from external data sources to help Large Language Models answer questions~\citep{borgeaud2022improving,gutiérrez2024hipporag,gutiérrez2025ragmemorynonparametriccontinual,packer2023memgpt,han2024retrieval,shi2024replug,yasunaga2023retrieval}.
However, none of the above has studied how to build a long-term memory system that could learn from both visual observations of the environment and social interactions with other agents, which we tackled with a dual-form structured memory and foundation models.

\begin{figure*}[t]
    \centering
\includegraphics[width=1.0\linewidth]{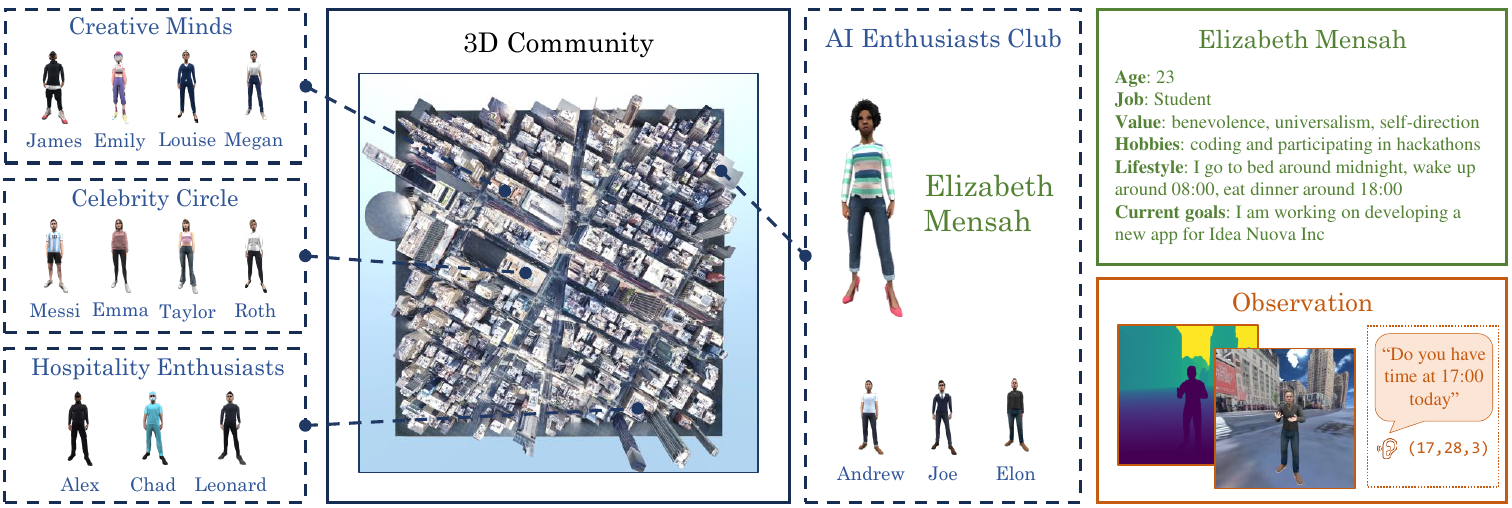}
    \caption{\textbf{An example community of 15 agents and 4 social groups in New York.} The character and observation of agent \textit{Elizabeth Mensah} are shown on the right.}
    \label{fig:setting}
\end{figure*}

\section{Problem Setting}

In our setting, $n$ agents with unique visual appearance $v_i$ and character profile $c_i$ inhabit an open, socially interactive world $W$, forming $k$ social groups, as illustrated in Figure~\ref{fig:setting}. 
Each character is defined by basic attributes such as name, age, occupation, values~\citep{schwartz2012overview}, hobbies, lifestyle, and current goals within the community. These attributes guide the agent’s daily decision-making. Social groups consist of a subset of agents selected based on character compatibility, and are defined by a group name, a detailed textual description, and a designated physical location for group activities. These groups connect the agents into a cohesive community, allowing rich and complex social interactions grounded in the 3D environment. 
Each agent is initialized with partial knowledge about the world, including known places and familiar agents, such as their residence and fellow group members, based on their character. 
The simulation runs at a fine temporal resolution of one second per step, during which each agent receives an observation $o_i$ including posed RGB and depth images, as well as dialogue content from nearby agents. Communication is spatial-constrained: agents can only engage in conversation if they are within a threshold distance $\theta_s$, mimicking realistic spatial constraints on verbal interactions. Every second, agents execute an action $a_i$ which may involve interacting with the environment or other agents.
During controlled evaluations, intervention occurs solely through modifications to agents’ community goals. Agents are required to make optimal decisions $a_i$ based on their updated character profiles  $c_i$ and incoming observations $o_i$.

\section{Ella: Embodied Lifelong Learning Agent}
\label{sec:method}
\vspace{-2mm}
To enable the embodied agents to continually learn within a community in a 3D open world, robust and efficient long-term memory is the key. Borrowing the concepts from psychology and cognitive neuroscience~\citep{tulving1972episodic}, we build long-term memory in two forms: name-centric semantic memory (Section~\ref{sec:smem}) and spatiotemporal episodic memory (Section~\ref{sec:emem}). Then in ~Section~\ref{sec:planning}, we introduce how we leverage the foundation models to integrate this memory system to facilitate the agent's everyday planning and social interactions.

\subsection{Name-centric Semantic Memory}
\label{sec:smem}
\vspace{-2mm}
Semantic memory stores facts about the agent and world, which is continually updated while the agent interacts with the world and other agents. Different from language agents, which normally take external databases like Wikipedia as a form of knowledge to help reasoning~\citep{sumers2023cognitive,lewis2020retrieval,borgeaud2022improving}, embodied agents need knowledge grounded in the environment they inhabit. We organize the different types of knowledge in a name-centric way and connect the related ones into a graph as shown in Figure~\ref{fig:framework} (a). Specifically, we build a hierarchical scene graph on the fly to serve as a spatial memory to help the agent navigate the visual world. The semantic memory is updated whenever there is a new visual observation made or a conversation finished, as introduced in Section~\ref{sec:comm}.

\subsubsection{Hierarchical Scene Graph as Spatial Memory}
\label{sec:sg}
\begin{figure}[t]
    \centering
\includegraphics[width=1.0\linewidth]{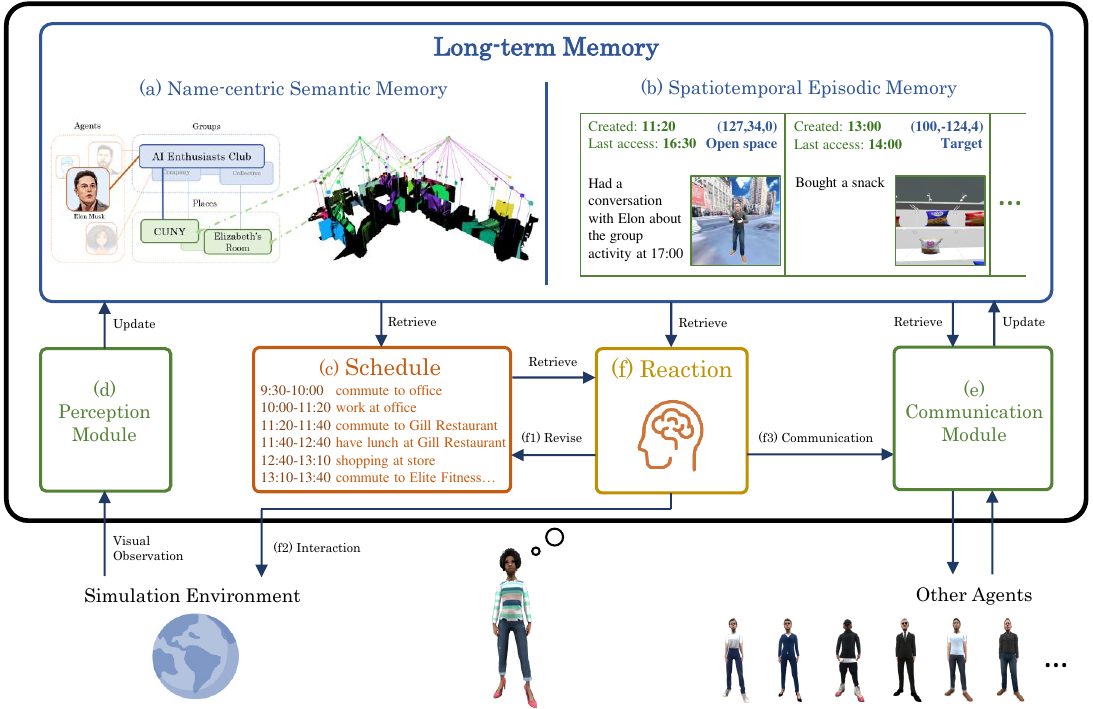}
    \caption{\textbf{Method Overview.} We build long-term memory in two forms: (a) name-centric semantic memory organizes the knowledge in a name-centric graph including a hierarchical scene graph serving as the spatial memory; (b) spatiotemporal episodic memory stores the experience as a series of events consisting of \textcolor{MyDarkGreen}{time}, \textcolor{MyDarkBlue}{location}, and multimodal contents. (c) \textit{Ella} first generates a daily schedule according to the knowledge and experiences retrieved from the long-term memory, (d) then updates the memory based on visual observations of the environment, and (e) social interactions with other agents and (f) makes reactions accordingly including (f1) revising the schedule, (f2) interacting with the environment, (f3) and engaging in a conversation.}
    \label{fig:framework}
    \vspace{-2mm}
\end{figure}

\vspace{-2mm}
Maintaining a spatial memory of the surrounding world is vital for embodied agents to act in a 3D world. To serve this purpose, we incrementally build a hierarchical scene graph~\citep{hughes2022hydra,gu2024conceptgraphs} on the fly as shown in Figure~\ref{fig:framework} (a). 

\noindent\textbf{Volume Grid Layer} Given posed RGB and depth observation, we first project them to 3D space and represent them in volume grid representations to act as low-level geometric memory. We then obtain an occupancy map based on it to facilitate navigation while avoiding obstacles in the 3D world. We divided the entire map into blocks of 0.5m $\times$ 0.5m and subdivided each block into smaller cells of 0.1m $\times$ 0.1m. We identified the lowest position within each small cell that could accommodate a person. A cell was classified as containing an obstacle if the height difference between this position and any of its neighboring cells exceeded 0.5m.

\noindent\textbf{Object Layer} Taking inspiration from previous works~\citep{gu2024conceptgraphs, maggio2024clio}, we employ a multi-stage perception pipeline to process RGB observations in an open world. Specifically, we utilize a combination of open-set vision models—including tagging~\citep{huang2023open}, object detection~\citep{liu2023grounding}, and segmentation~\citep{ravi2024sam2segmentimages}—to form the perception module. This module extracts a sequence of semantically labeled masks \(\langle m_i, \text{tag}_i \rangle\) as object candidates. 
Using depth and pose observations, each mask \(m_i\) is projected into a 3D point cloud \(p_i\), enabling the computation of geometric similarity \( \text{sim}(p_i, p_j) \) between objects based on their spatial overlap. Additionally, we extract visual features \(v_i\) for each object by encoding the corresponding cropped image~\citep{radford2021learning}. The detected object candidates from the current frame are then merged with existing objects based on similarity measurements.
Unlike \citep{gu2024conceptgraphs}, we handle the additional complexity of dynamic objects such as agents and vehicles. Due to the relatively low perception rate (1 FPS), conventional tracking techniques are impractical. Instead, we rely on visual similarity to associate and merge dynamic objects across frames.

\noindent\textbf{Region Layer}
We also implemented a region layer to further classify the buildings. First, we used the occupancy map and a breadth-first search to compute the Generalized Voronoi Diagram (GVD)\cite{hughes2022hydra} of the map. For each point \( p \) in the GVD, we determined the set \( S = \arg \min \{ \text{dist}(p,b) | b \in B \} \), where \( B \) represents the set of all buildings. We then connected all buildings in \( S \) with edges weighted by \( \frac{1}{\text{dist}(p,s)^2} \), where \( s \in S \). Finally, we connected all previously unconnected buildings by adding edges with zero weight, resulting in a complete graph. To group nodes connected by higher-weight edges, we applied spectral clustering, partitioning the graph into \( \sqrt{|B|} \) regions. This clustering facilitated a more structured geometric partitioning of the buildings.

\vspace{-4mm}
\subsection{Spatiotemporal Episodic Memory}
\label{sec:emem}
\vspace{-2mm}
Episodic memory is responsible for storing personal experiences~\citep{tulving1972episodic, tulving1983elements, lindes2016toward}. As noted by~\citep{mastrogiuseppe2019spatiotemporal}, episodic memory encodes not only when and what events occurred but also where they took place—highlighting the crucial role of spatial information. Unlike~\citep{park2023generative}, our episodic memory module incorporates both temporal and spatial information, in addition to multi-modal content, enabling the agent to retrieve experiences relevant to its current location. Experiences are stored as a series of events, each composed of temporal attributes (event creation time and last access time), spatial attributes (event location and place), and content attributes (a textual description and a corresponding egocentric image), as illustrated in Figure~\ref{fig:framework}(b).

\noindent\textbf{Retrieval} The episodic memory supports spatiotemporal retrieval. Given a query—comprising time, location, and content—all stored experience items are ranked based on the following three criteria:

\textit{Spatial Proximity} measures the distance between the event location $p_e$ and the query location $p_q$. $\text{proximity}(e, q) = \frac{1}{\|\mathbf{p}_e - \mathbf{p}_q\| + \epsilon}$

\textit{Content Relevance} measures how well an event's content aligns with the given query by evaluating both textual and visual similarity. Specifically, we compute the cosine similarity between the encoded representations of the event and query, considering both their text descriptions $T$ and images $I$. The final relevance score is obtained by averaging these two similarities. $\text{Relevance}(e,q) = (cos(T_e,T_q) + cos(I_e, I_q)) / 2$

\textit{Temporal Recency} is higher for events recently accessed. Following~\citep{park2023generative}, we model recency using an exponential decay function based on the time elapsed since the memory was last accessed. $\text{Recency}(e) = \exp\left(t_e-t_q\right)$

All three scores are then normalized to the range of $[0,1]$ with min-max scaling and averaged as the final score, and the top $k$ events are retrieved.

\vspace{-2mm}
\subsection{Planning, Reaction, and Communication}
\label{sec:planning}
\vspace{-2mm}
With this structured long-term memory, \textit{Ella} leverages foundation models to make efficient and robust everyday planning. Following \citep{park2023generative}, we adopt a planning and reaction framework with several modifications to facilitate efficient daily planning. \textit{Ella} first generates an environment- and characters-grounded daily schedule according to the knowledge and experiences retrieved from the long-term memory, then updates the memory based on observations and makes reactions accordingly, including revising the schedule, engaging in a conversation, and interacting with the environment. A specific communication module is incorporated to generate the utterance to chat about, summarize the conversations, and extract knowledge from it. More details on the submodules are provided in Appendix~\ref{app:method}. All prompt templates are provided in Appendix~\ref{app:prompt}.

\vspace{-2mm}
\subsubsection{Daily Schedule}
\vspace{-2mm}
At the start of each day, \textit{Ella} will retrieve experience and knowledge from the long-term memory with a query of \textit{``Things to consider for my schedule today.''}, then use foundation models to generate the daily schedule. Different from \citep{park2023generative}, we generate the daily schedule in a structured manner directly with each activity represented with a start time, an ending time, an activity description, and the corresponding activity place. Specifically, we consider the required commute time between adjacent activities happening in different places explicitly, due to the actual cost of navigating in an expansive 3D environment. For example, commuting from the office to a party place may take more than 15 minutes on foot, without considering that the agent may miss the party if they planned to attend the party at the party starting time. Figure~\ref{fig:framework} (c) shows a generated daily schedule for agent \textit{Elisabeth Mansah}. The daily schedule may be revised later by the reaction module given new experience and knowledge obtained from observations and social interactions during the day.

\vspace{-2mm}
\subsubsection{Reaction}
\vspace{-2mm}
Upon receiving new observations, the agent first processes visual information and updates its semantic memory using the perception module introduced in Section~\ref{sec:sg}. If new objects are detected or messages are heard, the agent invokes the reaction module. This module begins by retrieving relevant memories using the query \textit{``Important things to react to.''}, then use foundation models to reason about the character, current time, place, schedule, and retrieved memory and make one of the four choices: \textit{revising the schedule, interacting with the environment, engaging in a conversation, or no reactions needed}, as illustrated in Figure~\ref{fig:framework} (f). Additionally, the reaction module is automatically triggered if the time elapsed since the last reaction exceeds $\theta_{react}$ seconds.

\vspace{-2mm}
\subsubsection{Communication}
\label{sec:comm}
\vspace{-2mm}
When the agent generates a reaction of \textit{engaging in a conversation}, the communication module is revoked to generate the utterance to chat about by first retrieving the related knowledge and experience from the long-term memory with a query of the latest sentence in the conversation or \textit{``Things to chat about with \texttt{conversation targets}''} if the agent is initiating a new conversation, then use foundation models to synthesize the appropriate utterance.

When the conversation finishes, the communication module will summarize it and store the summarized conversation in episodic memory. \textit{Ella} will also try to extract new knowledge it learned from the conversation by prompting a foundation model with some demonstration knowledge items, and use it to update the semantic memory.

\section{Experiments}
\vspace{-2mm}
\input{table/main}
\subsection{Experimental Setup}
\label{sec:setup}
\vspace{-2mm}
We instantiate our embodied social agents community in \ textit {Virtual Community}~\citep{vico}, an open world simulation platform for multi-agent embodied AI, featuring large-scale community scenarios derived from the real world with realistic physics simulation and rendering based on Genesis~\citep{genesis}. We conducted experiments with 15 agents of unique characters in 3 different scenes and communities.  The observation includes posed $512\times512$ RGB and depth images, the content of the heard messages within range, and current states including pose, place, time, cash, held objects, and vehicles being taken. The agent's action space consists of navigation actions of 
\textit{move forward $x$ m}, \textit{turn left $x$ degree}, \textit{turn right $x$ degree}, \textit{enter $x$ place or vehicle}, and \textit{exit $x$ vehicle}; interaction actions of  \textit{pick $x$ object with hand $y$}, \textit{drop object in hand $x$}; and \textit{converse message $x$ with a range of $y$ m}. The message transmission range threshold $\theta_{msg}$ is set to 10m.

There are two stages of the experiment. In the first stage, 15 agents are simulated for 9 hours (34200 steps) for their first day in the community, during which the agents could familiarize themselves with the 600m * 600m scene and other agents and build memories. Then in the second stage, we test them with two controlled evaluations in the days following: \textbf{\textit{Influence Battle}} and \textbf{\textit{Leadership Quest}}. In \textbf{\textit{Influence Battle}}, two of the four groups will be asked to organize a party at a specific place in 6 hours, and the members need to go around the city, find and invite agents outside of their group to attend the party. This evaluation tests the agents' capability to impact other agents by persuading them to attend the parties, which requires the capability of social reasoning, persuasion, and decision-making. In \textbf{\textit{Leadership Quest}}, each of the four groups is assigned a task to purchase several items from various stores in the city and return within 3 hours. One member from each group is designated as the leader and is the only one given full details of the task, while the remaining members are simply instructed to assist the leader. This controlled evaluation setting challenges the agent’s leadership abilities, particularly in assigning sub-tasks based on the diverse personalities and resources of group members.

\noindent\textbf{Metrics} We evaluate agents' capability to influence others with \textit{show up rate}, the total number of agents showing up at any party organizing place during the 30-minute party time divided by the total number of agents; and \textit{the total number of conversations} the organizing parties engaged in, reflecting the efficiency of the invitations. In \textbf{\textit{Leadership Quest}}, we measure the success of agents' leadership and cooperation by \textit{average completion rate}, the number of fulfilled target items divided by the number of all target items averaged across all groups; 
and \textit{the total number of conversations} reflecting the efficacy of the communications among the agents.

\noindent\textbf{Baselines} To the best of our knowledge, there hasn't been any embodied social agent framework supporting social interaction within a community with open-world 3D scenes. The most related methods are CoELA~\citep{zhang2023building}, which only considered two agents within a constrained indoor scene for a specific task, and Generative Agents~\citep{park2023generative}, which assume oracle perception and use a predefined communication mechanism. We re-implemented these two methods in our setting as the baselines.
\vspace{-2mm}
\begin{itemize}[itemindent=2em,labelsep=3pt,labelwidth=1em,leftmargin=1em,itemsep=0em]
    \item \textit{CoELA}~\citep{zhang2023building} is a cooperative embodied agent framework. We replace their perception module with ours since there isn't a pretrained 2D segmentation model available under our open-world setting. We provide the character description to replace the CoELA's task-specific description.
    \item \textit{Generative Agent}~\citep{park2023generative} is a believable simulacrum of human behavior with an unimodality long-term memory. We adopt the perception module of Ella to convert visual observations into text descriptions and use the same occupancy map and a* algorithm for visual navigation.
\end{itemize}

\noindent\textbf{Implementation Details} For the perception module, we use open-set tagging model RAM++~\citep{huang2023open}, object detection model GroundingDINO~\citep{liu2023grounding}, and segmentation model SAM2~\citep{ravi2024sam2segmentimages}. For the embedding models, we use CLIP~\citep{radford2021learning} model \texttt{ViT-B-32-256} from openclip~\cite{ilharco_gabriel_2021_5143773} for images and \texttt{text-embedding-3-small} from Azure for text. We use \texttt{gpt-4o}\footnote{model version 2024-11-20} as the foundation model backbone for our method and \textit{CoELA}, and \texttt{gpt-35-turbo}\footnote{model version 0125} for \textit{Generative Agent}\footnote{We tried to implement Generative Agent with \texttt{gpt-4o}, but the original prompts broke often and it's too costly given its large quantity of API call}. We also test our method with open source foundation models \texttt{DeepSeek-R1-Distill-Qwen-14B} and \texttt{Qwen2.5-14B-Instruct} served with vLLM~\citep{kwon2023efficient} as the backbone in the experiments with oracle perception.

\subsection{Results}
\label{sec:results}
\begin{figure}[t]
    \centering
    \begin{subfigure}[t]{0.46\linewidth}
        \centering
        \includegraphics[width=\linewidth]{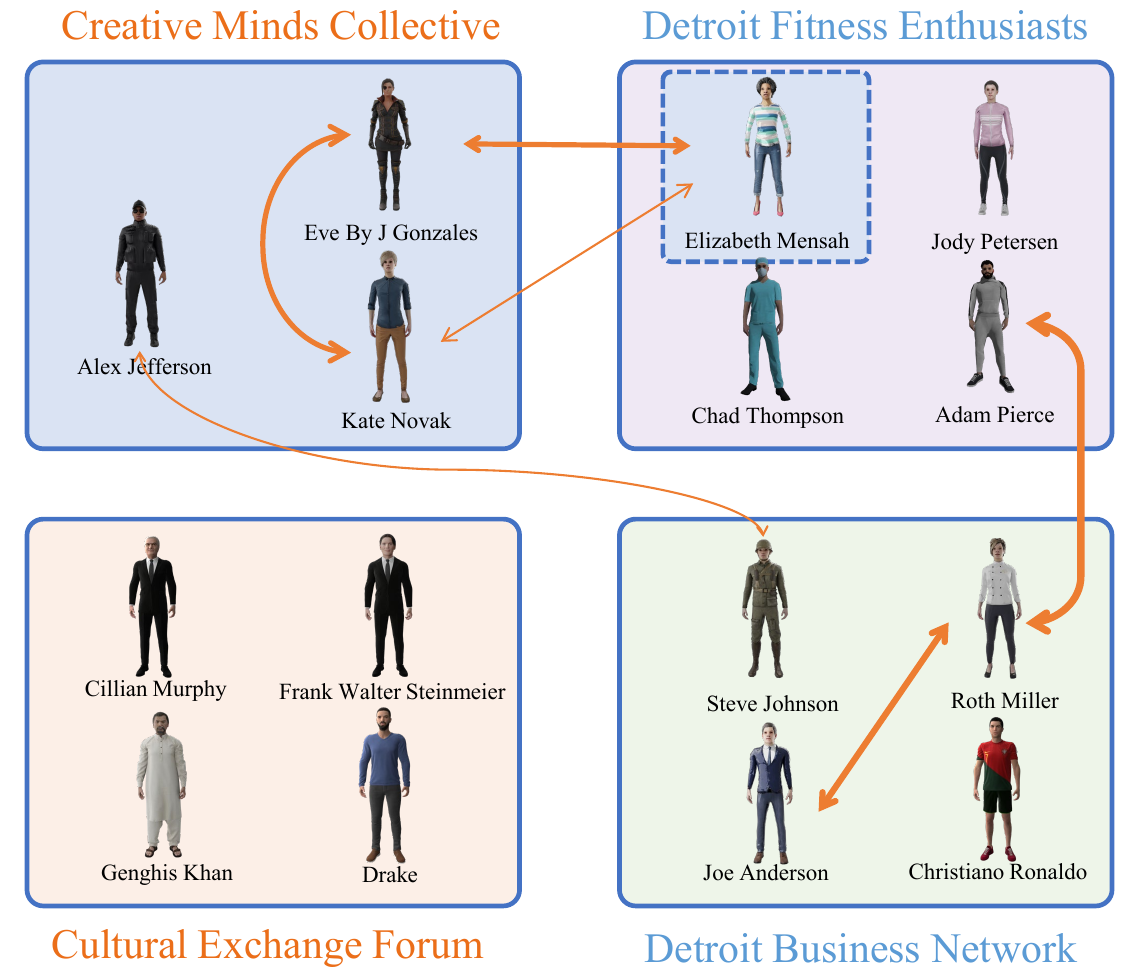}
        \caption{}
        \label{fig:social}
    \end{subfigure}
    \hfill
    \begin{subfigure}[t]{0.52\linewidth}
        \centering
        \includegraphics[width=\linewidth]{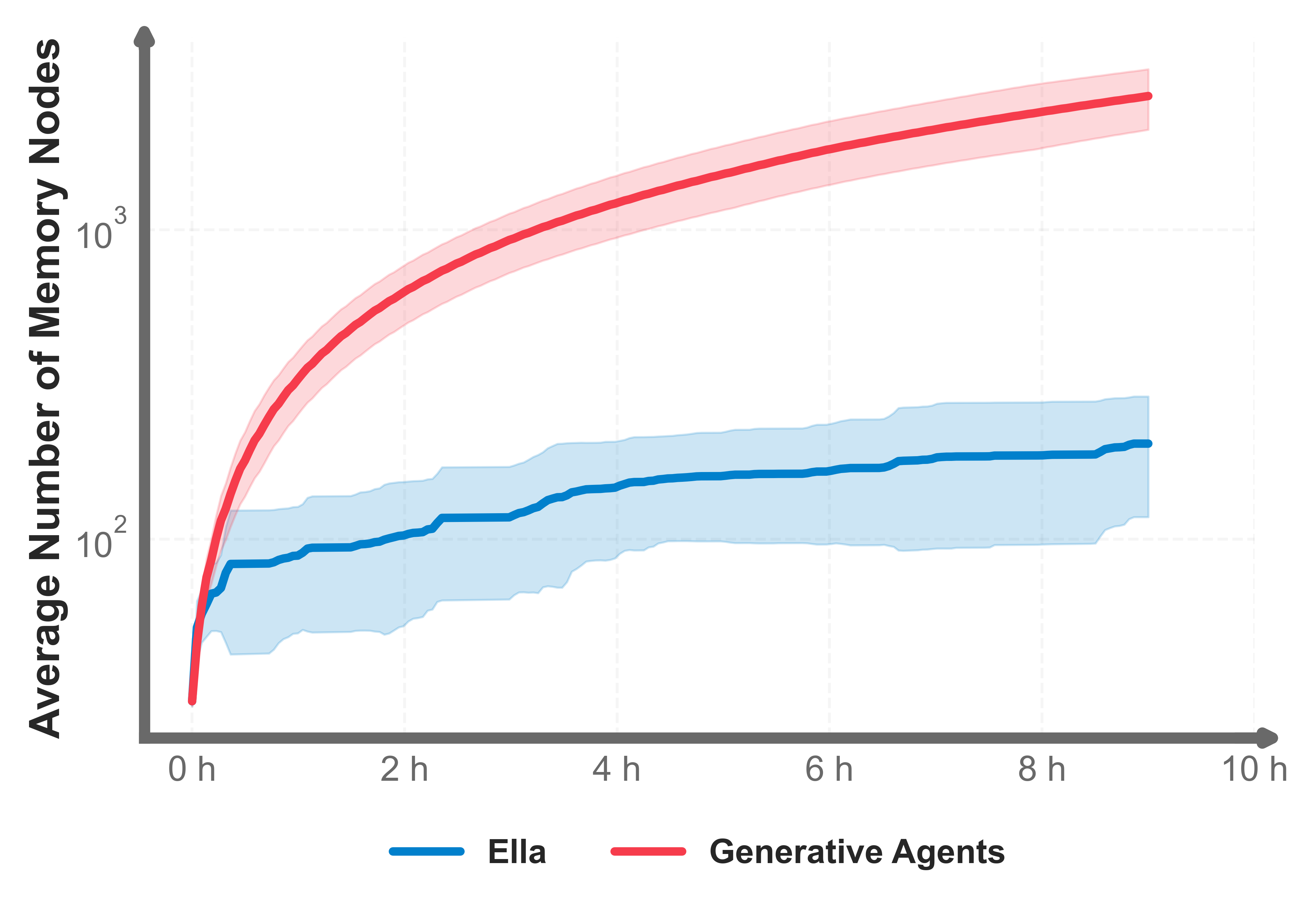}
        \caption{}
        \label{fig:memory_compare}
    \end{subfigure}
    \caption{(a) \textbf{Social interaction pattern in \textit{Influence Battle}.} The thickness of a line reflects the frequency of interaction. Members from Creative Minds Collective successfully persuaded Elizabeth Mensah to join their group's party. (b) \textbf{Comparison of memory growth over time.} The number of memory nodes averaged over 15 agents is shown here. Our structured memory system allows for more stable and organized growth.}
\end{figure}

\input{table/deepseek}

\noindent\textbf{\textit{Ella}'s structured long-term memory is efficient.} As shown in Figure~\ref{fig:memory_growth}, \textit{Ella} continuously accumulates new experiences and acquires new knowledge on the first day, covering nearly 50\% of the environment. An example of the final spatial coverage in the Detroit community is illustrated in Figure~\ref{fig:occ-map} in the Appendix. Figure~\ref{fig:memory_compare} further shows that \textit{Ella}'s structured memory system allows for more stable and organized growth of memory nodes compared to the Generative Agents baseline. This structure enables more efficient retrieval as memory scales, supporting timely access to relevant events even as the memory grows.

\noindent\textbf{\textit{Ella} can influence other agents effectively.} As shown in ~Table~\ref{tab:main}, \textit{Ella} achieves a higher show-up rate in the Influence Battle by successfully inviting more agents to the party across all three communities. This demonstrates its strong capabilities in social reasoning and persuasion. Although the \textit{CoELA} baseline engaged in twice as many conversations as \textit{Ella}, its show-up rate was only half as high. This discrepancy arises from its lack of long-term memory, preventing it from effectively recalling the party details after several hours (thousands of simulation steps). Meanwhile, Generative Agents engaged in so few conversations that they failed to invite other agents, despite being explicitly instructed to do so in their daily requirements. As illustrated in Figure~\ref{fig:social}, the party news propagates over time through the efforts of the organizer agents.

\noindent\textbf{\textit{Ella} can lead the group well.} As shown in Table~\ref{tab:main}, \textit{Ella} completes four times more goals than other baselines in the Leadership Quest. Notably, \textit{CoELA} had a completion rate of zero across all scenes—except in Detroit, where the leader partially completed the task alone—despite engaging in numerous conversations. This failure stems from its inability to retain memory of the required items. Among all scenarios, the London community posed the greatest challenge, where only \textit{Ella} achieved a non-zero performance, demonstrating the robustness of our approach.

\noindent\textbf{Robust perception is important for embodied social agents.} Different from \citep{park2023generative}'s setting where two agents knowing each other could only engage in a conversation when situated in the same grid, or \citep{zhang2023building}'s setting where two already-known agents could converse with each other anytime anywhere, our setting requires the agent to identify the agent to talk to according to their visual appearance or conversation contents and calculate the transmission range of their message according to the 3D location of the target agents to converse with, therefore a robust perception is critical for the agents to engage in social interactions in a 3D world. Comparing the results of \textit{Ella} and \textit{w/ Oracle Perception} in Table~\ref{tab:main}, we can see the performance further boots with Oracle perception, and there tend to be more conversations among the agents since they're more confident in identifying each other.

\noindent\textbf{Open source foundation models backbone is promising.} With the strong open-source foundation models like DeepSeek-R1~\citep{guo2025deepseek} becoming available recently, we wonder how well our framework works out-of-the-box on open-source foundation model backbones. We test \textit{Ella w/ Oracle Perception} Agent with different backbones across all three communities and the two controlled evaluations, the results are shown in Table~\ref{tab:deepseek}. Using \texttt{DeepSeek-R1-Distill-Qwen-14B} as the backbone without any further prompt engineering, \textit{Ella w/ Oracle Perception} achieves a reasonable performance close to that of using a backbone of \texttt{gpt-4o}, while \texttt{Qwen2.5-14B-Instruct} performs much worse.

\section{Limitations}
\label{sec:limitations}
\paragraph{Leverage the graph structure of the name-centric semantic memory.} Although the name-centric semantic memory is maintained as a graph structure, the current implementation retrieves knowledge based solely on text and image feature similarity. Enhancing our memory system with more sophisticated graph-based retrieval methods~\citep{zhang2025surveygraphretrievalaugmentedgeneration, sun2023think, gutiérrez2024hipporag} could enable effective multi-hop reasoning, paving the way for addressing reasoning-intensive challenges. This represents a promising direction for future work.

\paragraph{Lifelong simulation of a community of agents in a visually rich, physics-realistic environment is computationally expensive.} Although our experiments span only 1.5 simulated days—seemingly short for a "lifelong" setting—we adopt the widely used interpretation of lifelong learning as an agent’s ability to accumulate, retain, and reuse knowledge across experiences~\citep{chen2018lifelong}. Despite extensive system-level optimizations to accelerate simulation, each simulated second still requires at least one second of real time. This is due to the intensive demands of multi-camera rendering, skinned motion computation, and the invocation of multiple models or APIs during each agent’s decision-making process. As a result, simulating one day in the environment consumes an entire real-world day, significantly constraining the scale of experimentation. Continued progress in graphics and simulation technologies is expected to ease this bottleneck and support faster development of embodied social agents in high-fidelity, physics-grounded environments.

\paragraph{All agents' thinking processes are assumed to finish synchronously.} Human cognition is bounded by limited computational resources~\citep{lieder2020resource}. In our current setting, Agents are assumed to \textit{think} synchronously with unlimited computational resources, which means whatever deliberate the agent's thinking process is, it costs only 1 second in their world. It's interesting to consider the time cost of thinking given the same limited computational resources to all agents explicitly and study how agents could switch between slow system-2 thinking and fast system-1 thinking~\citep{evans2003two} adaptively.

\section{Conclusion}

In this work, we build a structured long-term memory with name-centric semantic memory and spatiotemporal episodic memory and introduce \textit{Ella}, an embodied social agent that uses foundation models and retrieved memory to reason, make daily plans, and engage in social activities. We conducted capability-oriented experiments in the Virtual Community with 15 agents in 3 different communities and demonstrated \textit{Ella} can use long-term memory effectively to influence, cooperate, and lead other agents in an open world while accumulating multi-modal experience and acquiring knowledge continuously from visual observations of the environment and social interactions with other agents. Our findings imply the power of combining structured long-term memory and foundation models to advance embodied general intelligence that could co-exist with humans.

\newpage
\newpage
\bibliographystyle{abbrv}
\bibliography{main}

\newpage

\newpage
\appendix

\section{Broader Impact}
\label{app:impacts}
As embodied social agents become more advanced, their integration into human-centered environments raises critical ethical and societal considerations. It's important to design and follow best practices in human-AI interactions~\citep{amershi2019guidelines}.

One key concern is the impact of AI-driven persuasion on human and agent interactions. In our \textit{\textbf{Influence Battle}} evaluation, \textit{Ella} successfully convinces other agents to attend an event, demonstrating its ability to shape group behavior. While such social reasoning capabilities are essential for cooperative AI, they could be misused in real-world applications, leading to manipulation, misinformation, or undue influence. To mitigate this, AI-driven persuasive agents must be designed with transparent intent disclosure and value alignment, ensuring they do not engage in deceptive or coercive behaviors.

Another concern is that their decision-making processes may inadvertently reflect and reinforce societal biases present in their training data or interaction patterns. For example, in our \textit{\textbf{Leadership Quest}}, Ella demonstrated superior leadership capabilities, but the fairness of leadership selection criteria in AI-driven systems remains an open question. Ensuring diversity and fairness in AI leadership roles requires robust bias mitigation strategies, careful dataset curation, and continuous evaluation of AI decision-making in diverse social contexts.

\input{99_appendix}

\end{document}

%% file: table/main.tex
\begin{table*}[t]
\centering
\vspace{-6mm}
\caption{\textbf{Main results.} We report the show-up rate and the total number of conversations for \textbf{Influence Battle}, and the completion rate and the total number of conversations for \textbf{Leadership Quest}. \textcolor{gray}{+ Oracle Perception} assumes ground truth 2D segmentation. The best results are in \textbf{bold}. \textbf{\textit{Ella}} achieves a higher show-up rate and completion rate across all three communities. }
\scalebox{0.78}{
	\begin{tabular}{lllllllll}
        \toprule
        & \multicolumn{4}{c}{\textbf{\textit{Influence Battle}}}
        & \multicolumn{4}{c}{\textbf{\textit{Leadership Quest}}}
        \\
        \cmidrule{2-5}\cmidrule{6-9}
        & New York & London & Detroit & \textbf{Average}  & New York & London & Detroit & \textbf{Average} \\
        \midrule
        \textbf{\textit{CoELA}~\citep{zhang2023building}} & 46.7, 57 & 20.0, 27 & 6.7, 17 & 24.5,	33.7 & 0.0, 72 & 0.0, 957 & 11.5, 625 & 3.8, 551.3 \\
        \textbf{\textit{Generative Agents}~\citep{park2023generative}} &  40.0, 3 & 40.0, 0 & 20.0, 0 & 33.3, 1.0 & 
        8.3, 169 & 0.0, 55 & 16.7, 14 & 8.3, 79.3\\
        \textbf{\textit{\textcolor{gray}{+ Oracle Perception}}} &  \textcolor{gray}{46.7, 5} & \textcolor{gray}{53.3, 153} & \textcolor{gray}{26.7, 0} & \textcolor{gray}{42.2, 52.7} & 
        \textcolor{gray}{4.2, 649} & \textcolor{gray}{0.0, 5} & \textcolor{gray}{16.7, 2} & \textcolor{gray}{7.0, 218.7}
        \\
        \textbf{\textit{Ella} (Ours)} & 46.7, 12 & 66.7, 19 & 46.7, 15 & \textbf{53.4, 15.3} & 
        33.3, 15 & 26.7, 17 & 37.5, 14 & \textbf{32.5, 15.3} \\
        \textbf{\textit{\textcolor{gray}{+ Oracle Perception}}} & \textcolor{gray}{60.0, 11} & \textcolor{gray}{60.0, 28} & \textcolor{gray}{53.3, 17} & \textcolor{gray}{\textbf{57.8, 18.7}} & 
        \textcolor{gray}{39.6, 87} & \textcolor{gray}{35.0, 35} & \textcolor{gray}{25.0, 26} & \textcolor{gray}{\textbf{33.2,	49.3}} \\
        \bottomrule
    \end{tabular}
}
    \vspace{-5mm}
    \label{tab:main}
\end{table*}

%% file: table/deepseek.tex
\begin{table}[t]
\centering
\caption{\textbf{Results with open-source foundation model backbone.} We report the results of \textit{Ella w/ Oracle Perception} with different backbones averaged over three communities here.}
\scalebox{0.9}{
\begin{tabular}{lcccc}
\toprule
& \multicolumn{2}{c}{\textbf{Influence Battle}} & \multicolumn{2}{c}{\textbf{Leadership Quest}} \\
\cmidrule{2-3} \cmidrule{4-5}
& \textbf{show-up rate} & \textbf{\# conversation} & \textbf{completion rate} & \textbf{\# conversation} \\
\midrule
\textbf{\textcolor{gray}{gpt-4o-1120}} & \textcolor{gray}{57.8} & \textcolor{gray}{18.7} & \textcolor{gray}{33.2} & \textcolor{gray}{49.3} \\
\textbf{DeepSeek-R1-Distill-Qwen-14B} & \textbf{40.0} & \textbf{48.3} & \textbf{8.0} & \textbf{46.0} \\
\textbf{Qwen2.5-14B-Instruct} & 22.2 & 89.0 & 1.0 & 57.3 \\
\bottomrule
\end{tabular}
}
\vspace{-5mm}
\label{tab:deepseek}
\end{table}

%% file: 99_appendix.tex
\section{Additional Experiment Details}

\subsection{Virtual Community}
\label{app:vico}

Virtual Community (ViCo), introduced by \citep{vico}, is an open world simulation platform for multi-agent embodied AI, featuring large-scale community scenarios derived from the real world with realistic physics and renderings. It was developed using Genesis~\citep{genesis} as its core engine, a generative physics simulator capable of modeling a wide variety of materials and an extensive array of robotic tasks, all while maintaining full differentiability. Additionally, Genesis features a real-time renderer based on OpenGL and a path-tracing renderer powered by Luisa. ViCo primarily offers scalable 3D scene creation and the generation of an embodied agent community.

\subsubsection{Scenes}
\begin{figure}[th]
\centering
\includegraphics[width=0.9\linewidth]{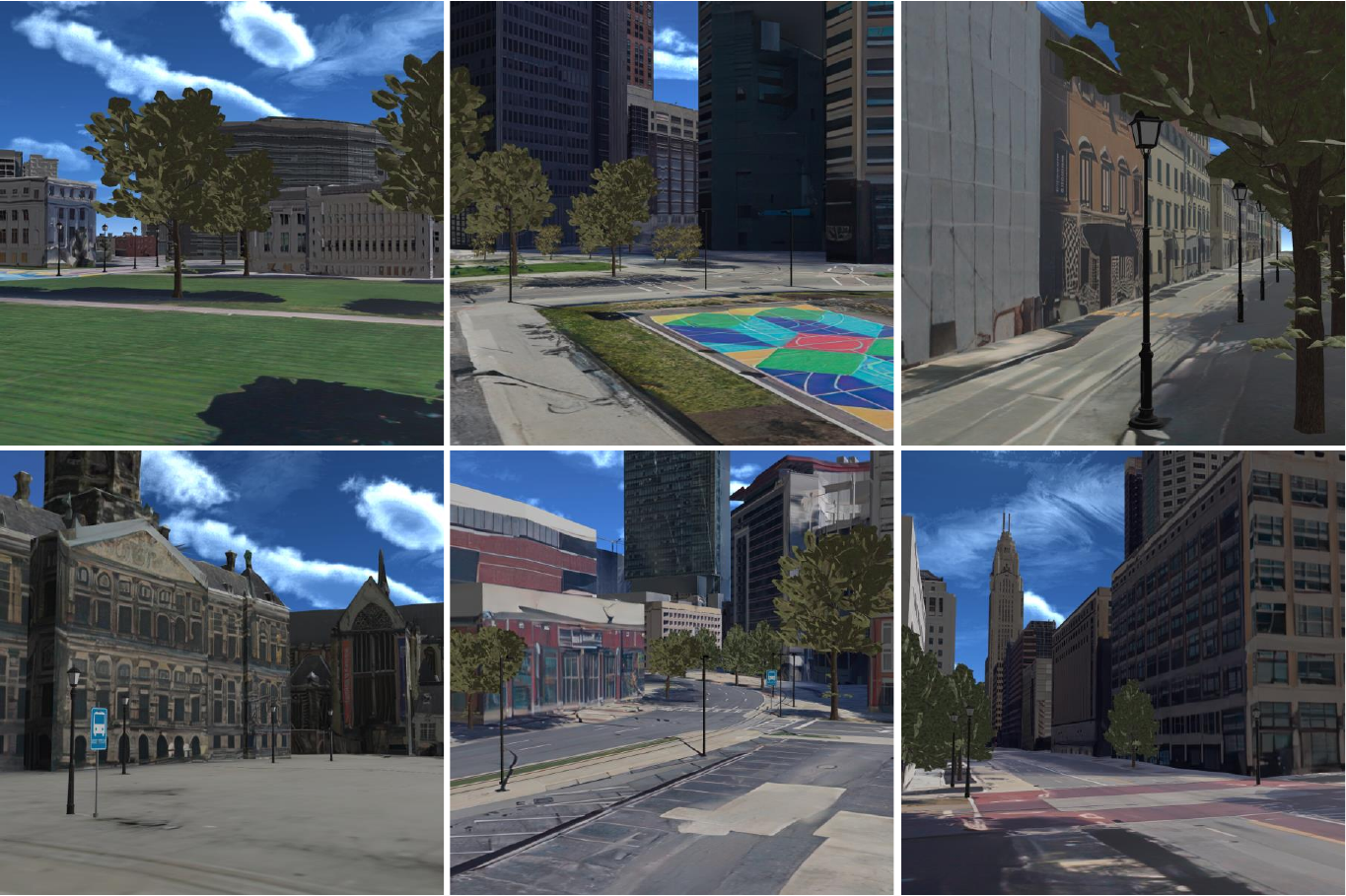}
\caption{\textbf{Close-up views of different scenes in Virtual Community.}}
\label{fig:vico-close}
\end{figure}

ViCo develops an online pipeline to transform existing 3D geospatial data into high-quality simulation-ready scenes. Moreover, the pipeline automatically annotates the scenes from these geospatial data to facilitate real-world alignment. It supports the creation of expansive outdoor and indoor environments at any location and scale. Currently, ViCo has generated 57 scenes of various cities worldwide. In this paper, we use a subset of 3 scenes from the generated scenes for our evaluation: New York City, Detroit, and London. Figure~\ref{fig:vico-close} presents views of different scenes within Virtual Community.

\subsubsection{Agents}
\begin{figure}[th]
    \centering
\includegraphics[width=1.0\linewidth]{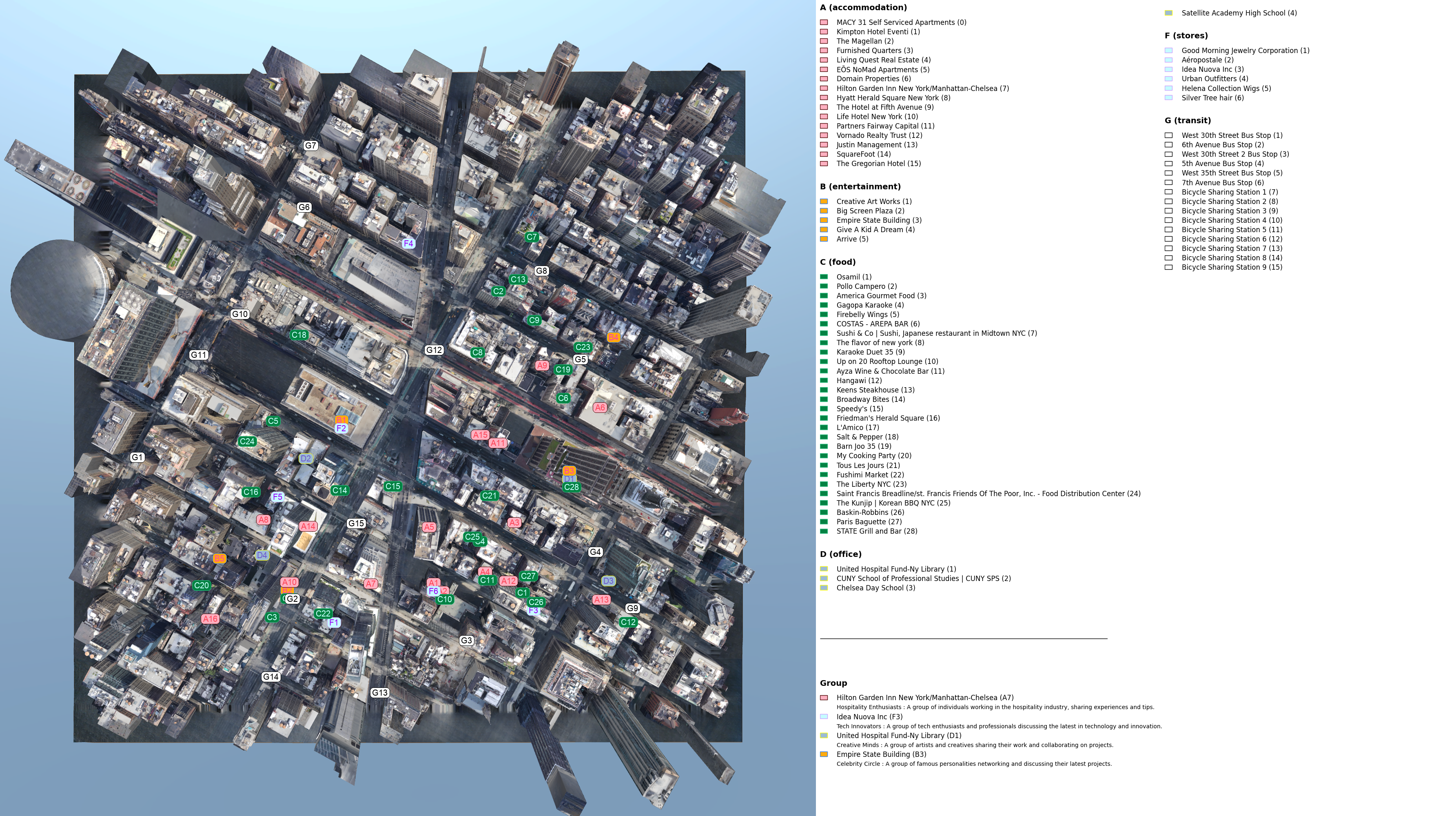}
    \caption{\textbf{An illustration of a community in New York City with places of different functionalities annotated.} There are 6 types of functional places: accommodation, entertainment, food, office, stores, and transit, each labeled with different colors on the figure. Social group information is also annotated with the group name, the group meeting place, and the group description.}
    \label{fig:known_places}
\end{figure}

ViCo has 74 avatar skins, consisting of ordinary skins from the Mixamo \footnote{\url{https://www.mixamo.com/}} and celebrity skins generated from real-world images using Avatar SDK \footnote{\url{https://avatarsdk.com}}. We randomly sampled 15 skins for each of the 3 scenes. ViCo combines SMPL-X human skeletons~\citep{SMPL-X:2019} with created avatar skins to support up to 2,299 unique motions from Mixamo. Additionally, ViCo can generate scene-grounded characters that are socially connected at a community level. Figure~\ref{fig:known_places} illustrates a generated community in New York City with places of different functionalities annotated.

\subsection{Compute}
\label{app:compute}
We conducted our experiments using a single NVIDIA A100 GPU. Stage one of each community life simulation was run for 20 hours, while stage two of each task and community was executed for an additional 10 hours. On average, each agent's saved memory—including both episodic and semantic components—occupies approximately 161 MB after 9 hours of simulation. During runtime, agents consume additional memory for perception, planning, and retrieval. In particular, the perception module alone requires around 4 GB of GPU memory per agent. The peak RAM usage per agent process is approximately 1 GB.

\section{Additional Implementation Details}
\label{app:method}

\subsection{Navigation}

\begin{figure}
    \centering
    \includegraphics[width=0.8\linewidth]{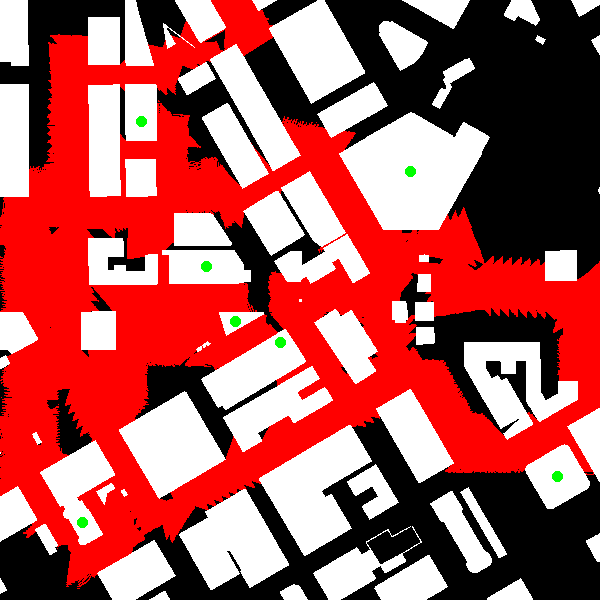}
    \caption{\textbf{A visualization of the final spatial coverage on the Detroit community.} Explored regions are shown in red, buildings are shown in white, and unexplored regions are shown in black. The buildings in the agent's schedule are denoted with green circles.}
    \label{fig:occ-map}
\end{figure}

Given the volume grid maintained in the semantic memory introduced in Section~\ref{sec:sg}, we construct the occupancy map and partition the entire map into three types of grid points: unknown, known obstacles, and known non-obstacles, as illustrated in Figure~\ref{fig:occ-map}. The A* algorithm is employed to search for the shortest path, where the weight of known non-obstacle points is set to 1, unknown points are assigned a weight of 5, and obstacle points are given an infinite weight. Additionally, to mitigate the issue of agents getting stuck near obstacles due to potential wall-clipping, points closer to obstacles are assigned higher weights. Specifically, a point at a distance \( d \) from an obstacle is assigned an additional weight of \( \frac{100}{d} \). Finally, to prevent the agent from wandering in place due to significant discrepancies between consecutive navigation paths, the previously computed path is prioritized unless it is found to be infeasible (i.e., it crosses an obstacle).

\section{Prompt Templates}
\label{app:prompt}

\begin{figure}[th]
    \centering
    \includegraphics[width=1.0\linewidth]{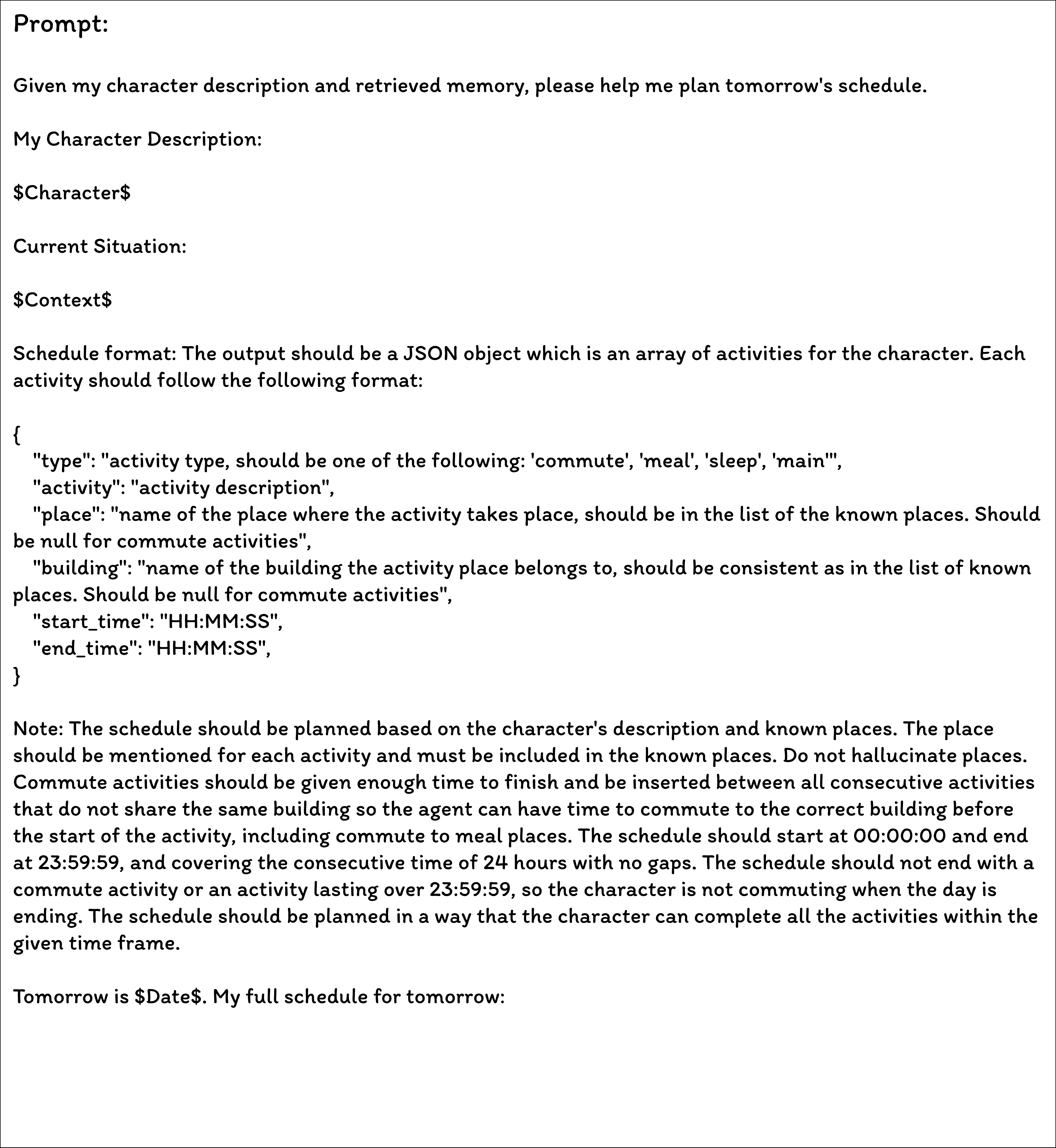}
    \caption{\textbf{Prompt template for generating the daily schedule.} \texttt{\$Character\$} is replaced with the agent's character description, \texttt{\$Context\$} is replaced with the retrieved memory.}
    \label{fig:plan-prompt}
\end{figure}

\begin{figure}[th]
    \centering
    \includegraphics[width=1.0\linewidth]{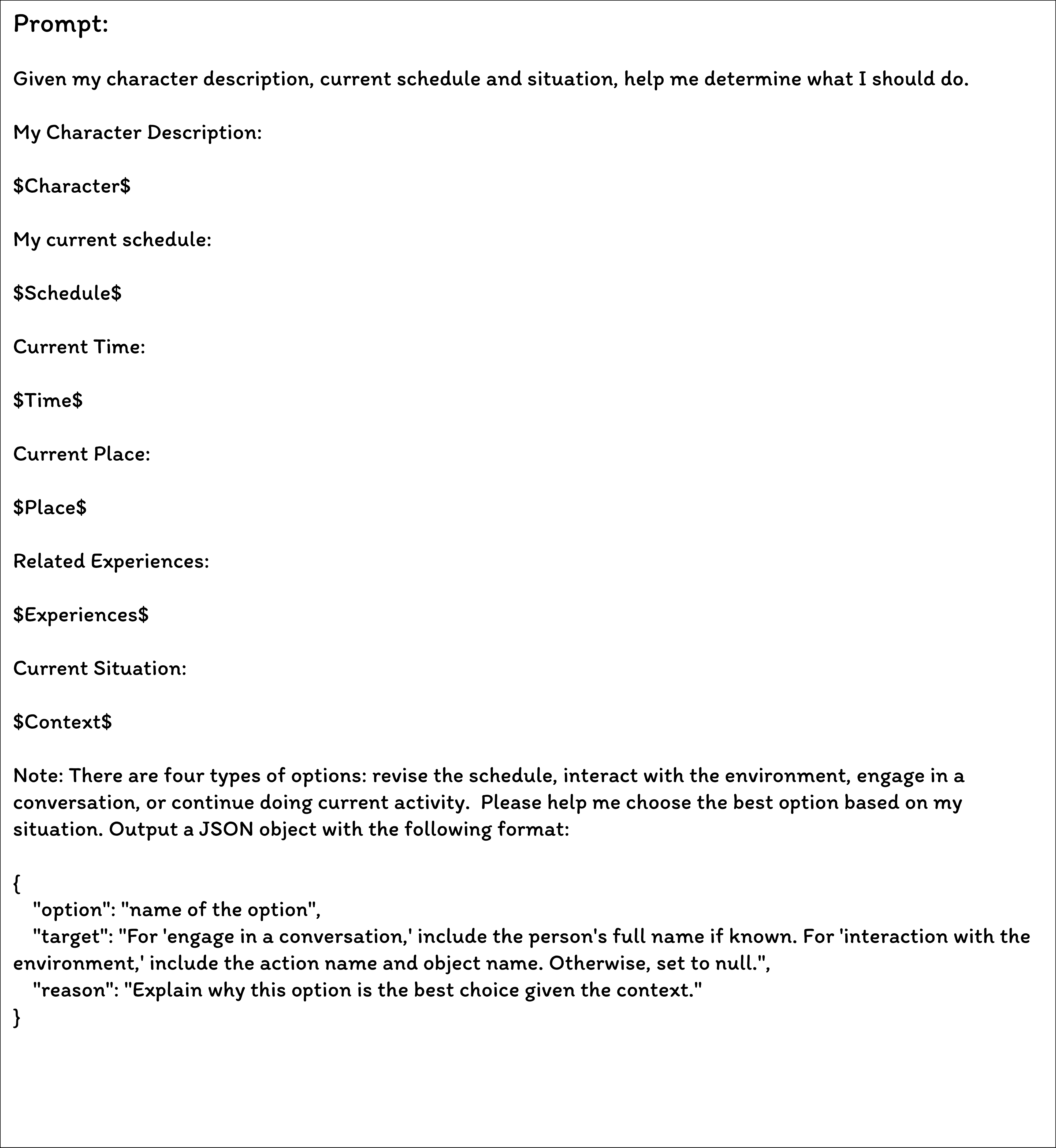}
    \caption{\textbf{Prompt template for generating the reaction.} \texttt{\$Character\$} is replaced with the agent's character description, \texttt{\$Schedule\$} is replaced with today's remaining schedules, \texttt{\$Experience\$} is replaced with the retrieved memory, \texttt{\$Context\$} is replaced with the latest memory.}
    \label{fig:reaction-prompt}
\end{figure}

\begin{figure}[th]
    \centering
    \includegraphics[width=1.0\linewidth]{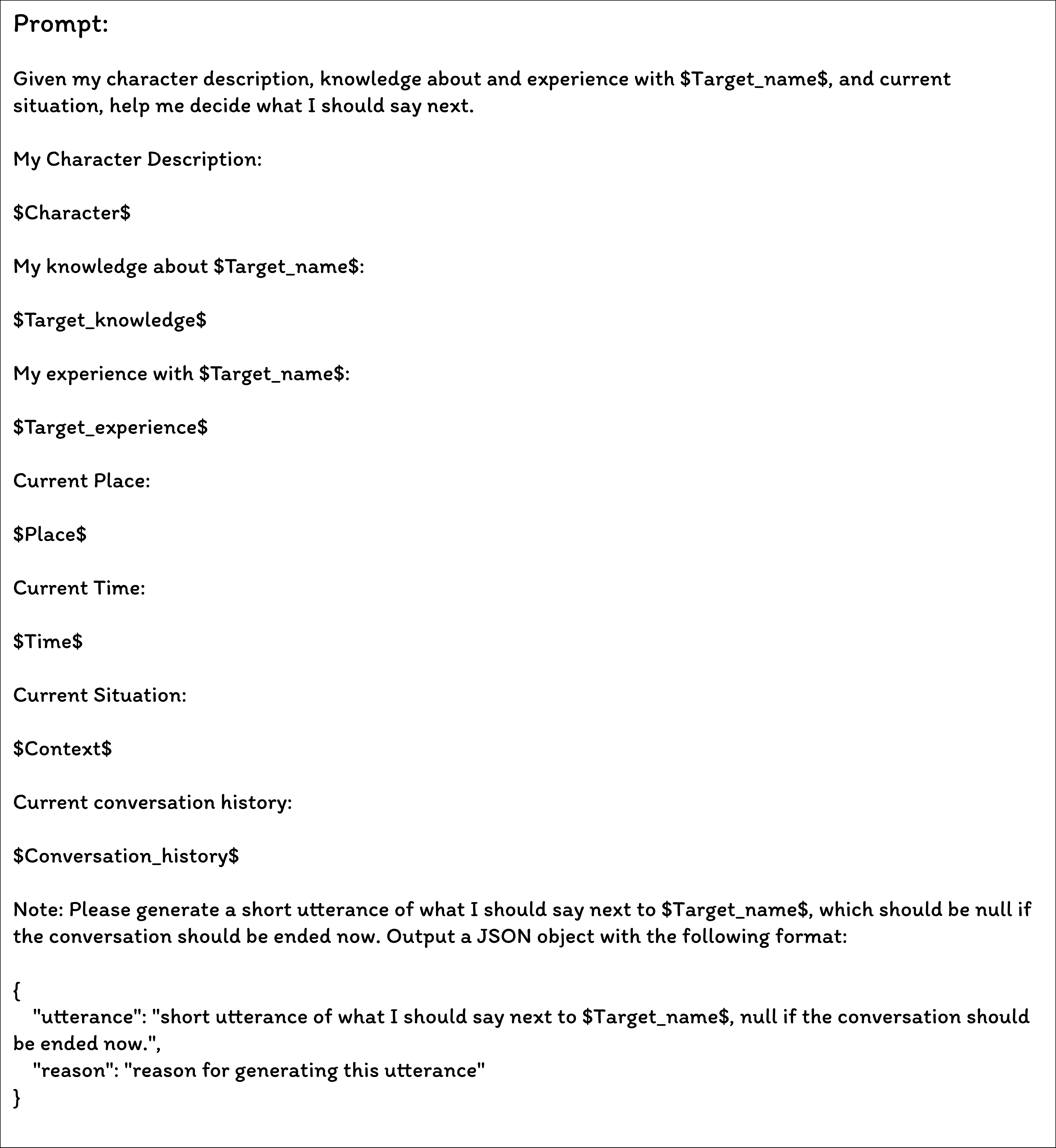}
    \caption{\textbf{Prompt template for generating the utterance.} \texttt{\$Character\$} is replaced with the agent's character description, \texttt{\$Target\_knowledge\$}, \texttt{\$Target\_experience\$}, \texttt{\$Context\$} are replaced with the retrieved memory, \texttt{\$Conversation\_history\$} is replaced with the last 4 messages.}
    \label{fig:utterance-prompt}
\end{figure}

\begin{figure}[th]
    \centering
    \includegraphics[width=1.0\linewidth]{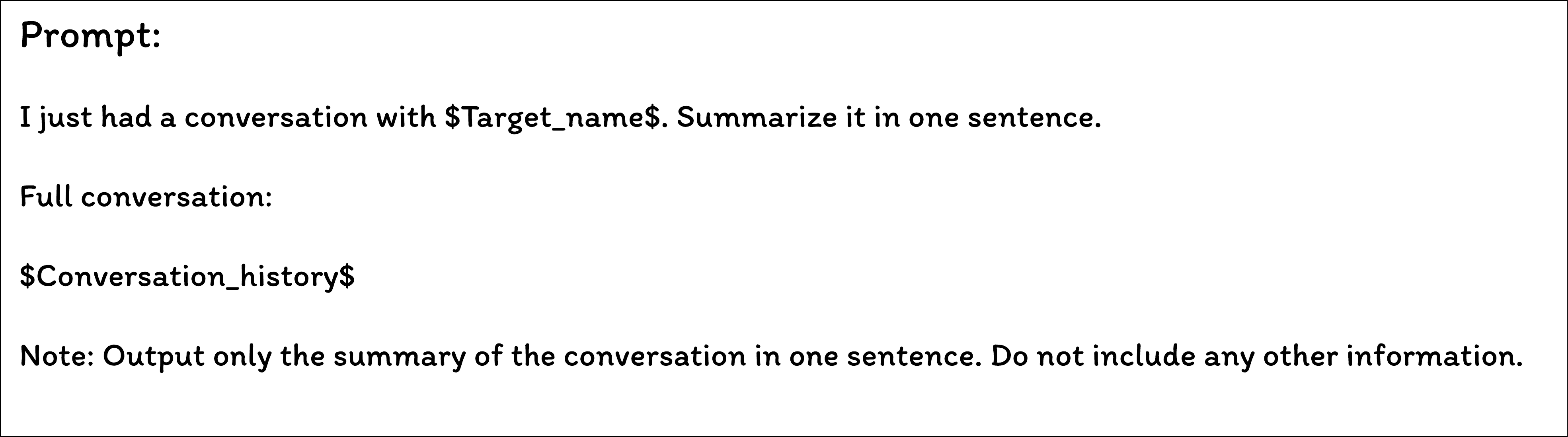}
    \caption{\textbf{Prompt template for generating the summarization of the conversation.} \texttt{\$Conversation\_history\$} is replaced with the full conversation.}
    \label{fig:summary-prompt}
\end{figure}

\begin{figure}[th]
    \centering
    \includegraphics[width=1.0\linewidth]{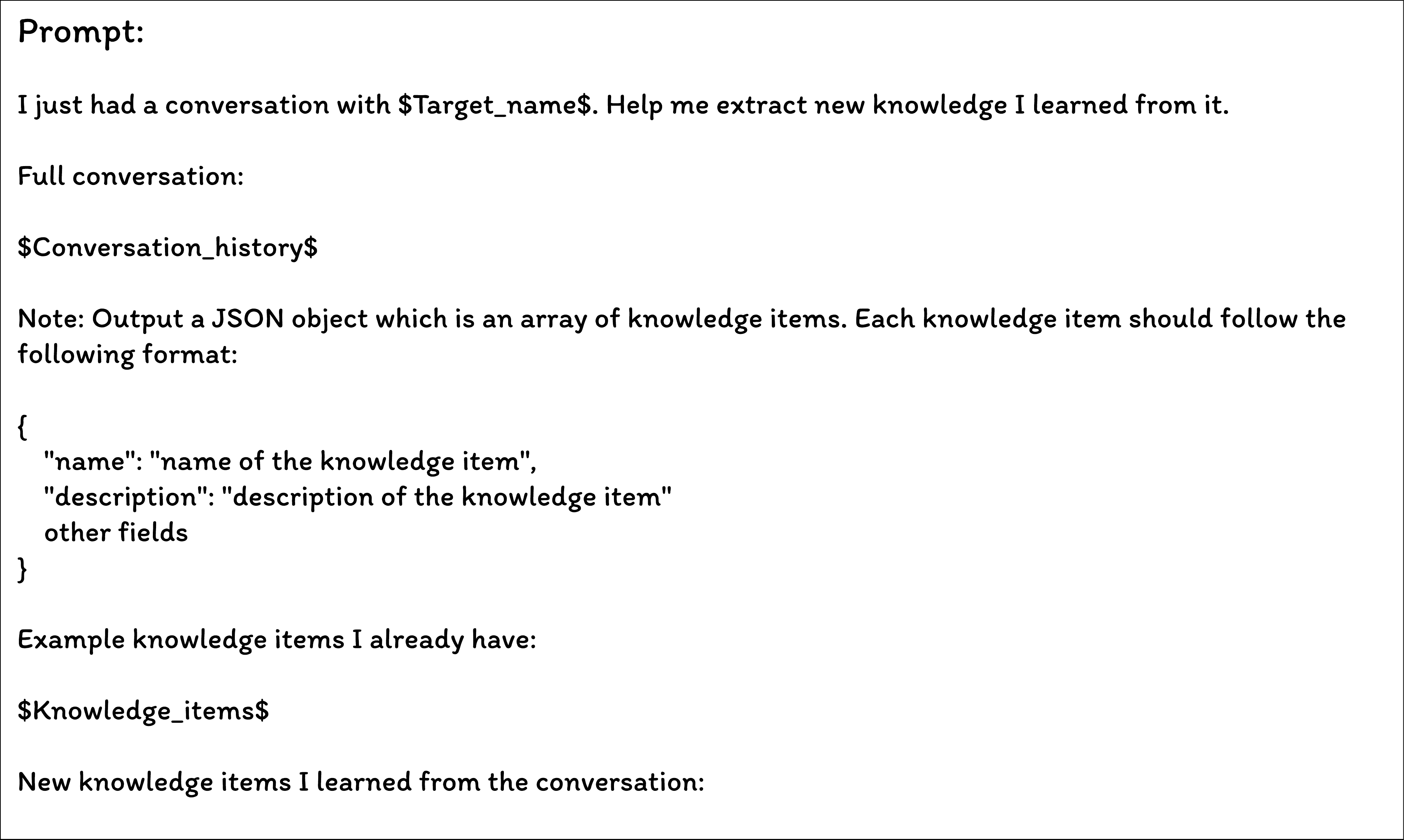}
    \caption{\textbf{Prompt template for extracting knowledge from a conversation.} \texttt{\$Conversation\_history\$} is replaced with the full conversation, \texttt{\$Knowledge\_items\$} is replaced with sampled knowledge items from semantic memory.}
    \label{fig:extract-prompt}
\end{figure}

We provide the full prompt template for the modules introduced in ~Section~\ref{sec:planning} in Figure~\ref{fig:plan-prompt} - Figure~\ref{fig:extract-prompt}.

%% file: neurips_2025.bbl
\begin{thebibliography}{100}

\bibitem{ahn2022can}
M.~Ahn, A.~Brohan, N.~Brown, Y.~Chebotar, O.~Cortes, B.~David, C.~Finn, K.~Gopalakrishnan, K.~Hausman, A.~Herzog, et~al.
\newblock Do as i can, not as i say: Grounding language in robotic affordances.
\newblock {\em arXiv preprint arXiv:2204.01691}, 2022.

\bibitem{amato2019modeling}
C.~Amato, G.~Konidaris, L.~P. Kaelbling, and J.~P. How.
\newblock Modeling and planning with macro-actions in decentralized pomdps.
\newblock {\em Journal of Artificial Intelligence Research}, 64:817--859, 2019.

\bibitem{amershi2019guidelines}
S.~Amershi, D.~Weld, M.~Vorvoreanu, A.~Fourney, B.~Nushi, P.~Collisson, J.~Suh, S.~Iqbal, P.~N. Bennett, K.~Inkpen, et~al.
\newblock Guidelines for human-ai interaction.
\newblock In {\em Proceedings of the 2019 chi conference on human factors in computing systems}, pages 1--13, 2019.

\bibitem{asgharivaskasi2023semantic}
A.~Asgharivaskasi and N.~Atanasov.
\newblock Semantic octree mapping and shannon mutual information computation for robot exploration.
\newblock {\em IEEE Transactions on Robotics}, 39(3):1910--1928, 2023.

\bibitem{genesis}
G.~Authors.
\newblock Genesis: A universal and generative physics engine for robotics and beyond, December 2024.

\bibitem{Baker2020Emergent}
B.~Baker, I.~Kanitscheider, T.~Markov, Y.~Wu, G.~Powell, B.~McGrew, and I.~Mordatch.
\newblock Emergent tool use from multi-agent autocurricula.
\newblock In {\em International Conference on Learning Representations}, 2020.

\bibitem{bard2020hanabi}
N.~Bard, J.~N. Foerster, S.~Chandar, N.~Burch, M.~Lanctot, H.~F. Song, E.~Parisotto, V.~Dumoulin, S.~Moitra, E.~Hughes, et~al.
\newblock The hanabi challenge: A new frontier for ai research.
\newblock {\em Artificial Intelligence}, 280:103216, 2020.

\bibitem{blukis2022persistent}
V.~Blukis, C.~Paxton, D.~Fox, A.~Garg, and Y.~Artzi.
\newblock A persistent spatial semantic representation for high-level natural language instruction execution.
\newblock In {\em Conference on Robot Learning}, pages 706--717. PMLR, 2022.

\bibitem{bobu2023aligning}
A.~Bobu, A.~Peng, P.~Agrawal, J.~Shah, and A.~D. Dragan.
\newblock Aligning robot and human representations.
\newblock {\em arXiv preprint arXiv:2302.01928}, 2023.

\bibitem{borgeaud2022improving}
S.~Borgeaud, A.~Mensch, J.~Hoffmann, T.~Cai, E.~Rutherford, K.~Millican, G.~B. Van Den~Driessche, J.-B. Lespiau, B.~Damoc, A.~Clark, et~al.
\newblock Improving language models by retrieving from trillions of tokens.
\newblock In {\em International conference on machine learning}, pages 2206--2240. PMLR, 2022.

\bibitem{carroll2019utility}
M.~Carroll, R.~Shah, M.~K. Ho, T.~Griffiths, S.~Seshia, P.~Abbeel, and A.~Dragan.
\newblock On the utility of learning about humans for human-ai coordination.
\newblock {\em Advances in neural information processing systems}, 32, 2019.

\bibitem{chaplot2020object}
D.~S. Chaplot, D.~P. Gandhi, A.~Gupta, and R.~R. Salakhutdinov.
\newblock Object goal navigation using goal-oriented semantic exploration.
\newblock {\em Advances in Neural Information Processing Systems}, 33:4247--4258, 2020.

\bibitem{chen2024socialbench}
H.~Chen, H.~Chen, M.~Yan, W.~Xu, G.~Xing, W.~Shen, X.~Quan, C.~Li, J.~Zhang, and F.~Huang.
\newblock Socialbench: Sociality evaluation of role-playing conversational agents.
\newblock In {\em Findings of the Association for Computational Linguistics ACL 2024}, pages 2108--2126, 2024.

\bibitem{chen2018lifelong}
Z.~Chen and B.~Liu.
\newblock {\em Lifelong machine learning}.
\newblock Morgan \& Claypool Publishers, 2018.

\bibitem{crosby2019animal}
M.~Crosby, B.~Beyret, and M.~Halina.
\newblock The animal-ai olympics.
\newblock {\em Nature Machine Intelligence}, 1(5):257--257, 2019.

\bibitem{dai2024artificial}
G.~Dai, W.~Zhang, J.~Li, S.~Yang, S.~Rao, A.~Caetano, M.~Sra, et~al.
\newblock Artificial leviathan: Exploring social evolution of llm agents through the lens of hobbesian social contract theory.
\newblock {\em arXiv preprint arXiv:2406.14373}, 2024.

\bibitem{dautenhahn2007socially}
K.~Dautenhahn.
\newblock Socially intelligent robots: dimensions of human--robot interaction.
\newblock {\em Philosophical transactions of the royal society B: Biological sciences}, 362(1480):679--704, 2007.

\bibitem{du2023video}
Y.~Du, M.~Yang, P.~Florence, F.~Xia, A.~Wahid, B.~Ichter, P.~Sermanet, T.~Yu, P.~Abbeel, J.~B. Tenenbaum, et~al.
\newblock Video language planning.
\newblock {\em arXiv preprint arXiv:2310.10625}, 2023.

\bibitem{evans2003two}
J.~S.~B. Evans.
\newblock In two minds: dual-process accounts of reasoning.
\newblock {\em Trends in cognitive sciences}, 7(10):454--459, 2003.

\bibitem{gadre2022continuous}
S.~Y. Gadre, K.~Ehsani, S.~Song, and R.~Mottaghi.
\newblock Continuous scene representations for embodied ai.
\newblock In {\em Proceedings of the IEEE/CVF Conference on Computer Vision and Pattern Recognition}, pages 14849--14859, 2022.

\bibitem{gadre2023cows}
S.~Y. Gadre, M.~Wortsman, G.~Ilharco, L.~Schmidt, and S.~Song.
\newblock Cows on pasture: Baselines and benchmarks for language-driven zero-shot object navigation.
\newblock In {\em Proceedings of the IEEE/CVF Conference on Computer Vision and Pattern Recognition}, pages 23171--23181, 2023.

\bibitem{gan2021threedworld}
C.~Gan, J.~Schwartz, S.~Alter, D.~Mrowca, M.~Schrimpf, J.~Traer, J.~D. Freitas, J.~Kubilius, A.~Bhandwaldar, N.~Haber, M.~Sano, K.~Kim, E.~Wang, M.~Lingelbach, A.~Curtis, K.~T. Feigelis, D.~Bear, D.~Gutfreund, D.~D. Cox, A.~Torralba, J.~J. DiCarlo, J.~B. Tenenbaum, J.~Mcdermott, and D.~L. Yamins.
\newblock Three{DW}orld: A platform for interactive multi-modal physical simulation.
\newblock In {\em Thirty-fifth Conference on Neural Information Processing Systems Datasets and Benchmarks Track (Round 1)}, 2021.

\bibitem{gombolay2015decision}
M.~C. Gombolay, R.~A. Gutierrez, S.~G. Clarke, G.~F. Sturla, and J.~A. Shah.
\newblock Decision-making authority, team efficiency and human worker satisfaction in mixed human--robot teams.
\newblock {\em Autonomous Robots}, 39:293--312, 2015.

\bibitem{goodrich2008human}
M.~A. Goodrich, A.~C. Schultz, et~al.
\newblock Human--robot interaction: a survey.
\newblock {\em Foundations and Trends{\textregistered} in Human--Computer Interaction}, 1(3):203--275, 2008.

\bibitem{gu2024conceptgraphs}
Q.~Gu, A.~Kuwajerwala, S.~Morin, K.~M. Jatavallabhula, B.~Sen, A.~Agarwal, C.~Rivera, W.~Paul, K.~Ellis, R.~Chellappa, et~al.
\newblock Conceptgraphs: Open-vocabulary 3d scene graphs for perception and planning.
\newblock In {\em 2024 IEEE International Conference on Robotics and Automation (ICRA)}, pages 5021--5028. IEEE, 2024.

\bibitem{guo2025deepseek}
D.~Guo, D.~Yang, H.~Zhang, J.~Song, R.~Zhang, R.~Xu, Q.~Zhu, S.~Ma, P.~Wang, X.~Bi, et~al.
\newblock Deepseek-r1: Incentivizing reasoning capability in llms via reinforcement learning.
\newblock {\em arXiv preprint arXiv:2501.12948}, 2025.

\bibitem{gur2023real}
I.~Gur, H.~Furuta, A.~Huang, M.~Safdari, Y.~Matsuo, D.~Eck, and A.~Faust.
\newblock A real-world webagent with planning, long context understanding, and program synthesis.
\newblock {\em arXiv preprint arXiv:2307.12856}, 2023.

\bibitem{gutiérrez2024hipporag}
B.~J. Gutiérrez, Y.~Shu, Y.~Gu, M.~Yasunaga, and Y.~Su.
\newblock Hipporag: Neurobiologically inspired long-term memory for large language models.
\newblock In {\em The Thirty-eighth Annual Conference on Neural Information Processing Systems}, 2024.

\bibitem{gutiérrez2025ragmemorynonparametriccontinual}
B.~J. Gutiérrez, Y.~Shu, W.~Qi, S.~Zhou, and Y.~Su.
\newblock From rag to memory: Non-parametric continual learning for large language models, 2025.

\bibitem{han2024retrieval}
H.~Han, Y.~Wang, H.~Shomer, K.~Guo, J.~Ding, Y.~Lei, M.~Halappanavar, R.~A. Rossi, S.~Mukherjee, X.~Tang, et~al.
\newblock Retrieval-augmented generation with graphs (graphrag).
\newblock {\em arXiv preprint arXiv:2501.00309}, 2024.

\bibitem{hong2024cogagent}
W.~Hong, W.~Wang, Q.~Lv, J.~Xu, W.~Yu, J.~Ji, Y.~Wang, Z.~Wang, Y.~Dong, M.~Ding, et~al.
\newblock Cogagent: A visual language model for gui agents.
\newblock In {\em Proceedings of the IEEE/CVF Conference on Computer Vision and Pattern Recognition}, pages 14281--14290, 2024.

\bibitem{hornung2013octomap}
A.~Hornung, K.~M. Wurm, M.~Bennewitz, C.~Stachniss, and W.~Burgard.
\newblock Octomap: An efficient probabilistic 3d mapping framework based on octrees.
\newblock {\em Autonomous robots}, 34:189--206, 2013.

\bibitem{huang2023visual}
C.~Huang, O.~Mees, A.~Zeng, and W.~Burgard.
\newblock Visual language maps for robot navigation.
\newblock In {\em 2023 IEEE International Conference on Robotics and Automation (ICRA)}, pages 10608--10615. IEEE, 2023.

\bibitem{huang2023voxposer}
W.~Huang, C.~Wang, R.~Zhang, Y.~Li, J.~Wu, and L.~Fei-Fei.
\newblock Voxposer: Composable 3d value maps for robotic manipulation with language models.
\newblock In {\em Conference on Robot Learning}, pages 540--562. PMLR, 2023.

\bibitem{huang2023open}
X.~Huang, Y.-J. Huang, Y.~Zhang, W.~Tian, R.~Feng, Y.~Zhang, Y.~Xie, Y.~Li, and L.~Zhang.
\newblock Open-set image tagging with multi-grained text supervision.
\newblock {\em arXiv e-prints}, pages arXiv--2310, 2023.

\bibitem{hughes2022hydra}
N.~Hughes, Y.~Chang, and L.~Carlone.
\newblock Hydra: A real-time spatial perception system for 3d scene graph construction and optimization.
\newblock {\em arXiv preprint arXiv:2201.13360}, 2022.

\bibitem{ilharco_gabriel_2021_5143773}
G.~Ilharco, M.~Wortsman, R.~Wightman, C.~Gordon, N.~Carlini, R.~Taori, A.~Dave, V.~Shankar, H.~Namkoong, J.~Miller, H.~Hajishirzi, A.~Farhadi, and L.~Schmidt.
\newblock Openclip, July 2021.
\newblock If you use this software, please cite it as below.

\bibitem{jaderberg2019human}
M.~Jaderberg, W.~M. Czarnecki, I.~Dunning, L.~Marris, G.~Lever, A.~G. Castaneda, C.~Beattie, N.~C. Rabinowitz, A.~S. Morcos, A.~Ruderman, et~al.
\newblock Human-level performance in 3d multiplayer games with population-based reinforcement learning.
\newblock {\em Science}, 364(6443):859--865, 2019.

\bibitem{jain2020cordial}
U.~Jain, L.~Weihs, E.~Kolve, A.~Farhadi, S.~Lazebnik, A.~Kembhavi, and A.~Schwing.
\newblock A cordial sync: Going beyond marginal policies for multi-agent embodied tasks.
\newblock In {\em Computer Vision--ECCV 2020: 16th European Conference, Glasgow, UK, August 23--28, 2020, Proceedings, Part V 16}, pages 471--490. Springer, 2020.

\bibitem{jiang2024long}
X.~Jiang, F.~Li, H.~Zhao, J.~Wang, J.~Shao, S.~Xu, S.~Zhang, W.~Chen, X.~Tang, Y.~Chen, et~al.
\newblock Long term memory: The foundation of ai self-evolution.
\newblock {\em arXiv preprint arXiv:2410.15665}, 2024.

\bibitem{kurenkov2023modeling}
A.~Kurenkov, M.~Lingelbach, T.~Agarwal, E.~Jin, C.~Li, R.~Zhang, L.~Fei-Fei, J.~Wu, S.~Savarese, and R.~Mart{\i}n-Mart{\i}n.
\newblock Modeling dynamic environments with scene graph memory.
\newblock In {\em International Conference on Machine Learning}, pages 17976--17993. PMLR, 2023.

\bibitem{kwon2023efficient}
W.~Kwon, Z.~Li, S.~Zhuang, Y.~Sheng, L.~Zheng, C.~H. Yu, J.~E. Gonzalez, H.~Zhang, and I.~Stoica.
\newblock Efficient memory management for large language model serving with pagedattention.
\newblock In {\em Proceedings of the ACM SIGOPS 29th Symposium on Operating Systems Principles}, 2023.

\bibitem{laird2022introduction}
J.~E. Laird.
\newblock Introduction to soar.
\newblock {\em arXiv preprint arXiv:2205.03854}, 2022.

\bibitem{lasota2017survey}
P.~A. Lasota, T.~Fong, J.~A. Shah, et~al.
\newblock A survey of methods for safe human-robot interaction.
\newblock {\em Foundations and Trends{\textregistered} in Robotics}, 5(4):261--349, 2017.

\bibitem{lewis2020retrieval}
P.~Lewis, E.~Perez, A.~Piktus, F.~Petroni, V.~Karpukhin, N.~Goyal, H.~K{\"u}ttler, M.~Lewis, W.-t. Yih, T.~Rockt{\"a}schel, et~al.
\newblock Retrieval-augmented generation for knowledge-intensive nlp tasks.
\newblock {\em Advances in Neural Information Processing Systems}, 33:9459--9474, 2020.

\bibitem{li2023behavior}
C.~Li, R.~Zhang, J.~Wong, C.~Gokmen, S.~Srivastava, R.~Mart{\'\i}n-Mart{\'\i}n, C.~Wang, G.~Levine, M.~Lingelbach, J.~Sun, et~al.
\newblock Behavior-1k: A benchmark for embodied ai with 1,000 everyday activities and realistic simulation.
\newblock In {\em Conference on Robot Learning}, pages 80--93. PMLR, 2023.

\bibitem{li2023camel}
G.~Li, H.~Hammoud, H.~Itani, D.~Khizbullin, and B.~Ghanem.
\newblock Camel: Communicative agents for" mind" exploration of large language model society.
\newblock {\em Advances in Neural Information Processing Systems}, 36:51991--52008, 2023.

\bibitem{li2019robust}
S.~Li, Y.~Wu, X.~Cui, H.~Dong, F.~Fang, and S.~Russell.
\newblock Robust multi-agent reinforcement learning via minimax deep deterministic policy gradient.
\newblock In {\em Proceedings of the AAAI conference on artificial intelligence}, volume~33, pages 4213--4220, 2019.

\bibitem{li2022embodied}
X.~Li, D.~Guo, H.~Liu, and F.~Sun.
\newblock Embodied semantic scene graph generation.
\newblock In {\em Conference on robot learning}, pages 1585--1594. PMLR, 2022.

\bibitem{li2024optimus}
Z.~Li, Y.~Xie, R.~Shao, G.~Chen, D.~Jiang, and L.~Nie.
\newblock Optimus-1: Hybrid multimodal memory empowered agents excel in long-horizon tasks.
\newblock In {\em The Thirty-eighth Annual Conference on Neural Information Processing Systems}, 2024.

\bibitem{lieder2020resource}
F.~Lieder and T.~L. Griffiths.
\newblock Resource-rational analysis: Understanding human cognition as the optimal use of limited computational resources.
\newblock {\em Behavioral and brain sciences}, 43:e1, 2020.

\bibitem{lindes2016toward}
P.~Lindes and J.~E. Laird.
\newblock Toward integrating cognitive linguistics and cognitive language processing.
\newblock In {\em Proceedings of the 14th International Conference on Cognitive Modeling (ICCM)}, 2016.

\bibitem{liu2024training}
R.~Liu, R.~Yang, C.~Jia, G.~Zhang, D.~Yang, and S.~Vosoughi.
\newblock Training socially aligned language models on simulated social interactions.
\newblock In {\em The Twelfth International Conference on Learning Representations}, 2024.

\bibitem{liu2023grounding}
S.~Liu, Z.~Zeng, T.~Ren, F.~Li, H.~Zhang, J.~Yang, Q.~Jiang, C.~Li, J.~Yang, H.~Su, et~al.
\newblock Grounding dino: Marrying dino with grounded pre-training for open-set object detection.
\newblock {\em arXiv preprint arXiv:2303.05499}, 2023.

\bibitem{liu2024exploring}
X.~Liu, J.~Zhang, S.~Guo, H.~Shang, C.~Yang, and Q.~Zhu.
\newblock Exploring prosocial irrationality for llm agents: A social cognition view.
\newblock {\em arXiv preprint arXiv:2405.14744}, 2024.

\bibitem{liu2024interintent}
Z.~Liu, A.~Anand, P.~Zhou, J.-t. Huang, and J.~Zhao.
\newblock Interintent: Investigating social intelligence of llms via intention understanding in an interactive game context.
\newblock {\em arXiv preprint arXiv:2406.12203}, 2024.

\bibitem{losey2022learning}
D.~P. Losey, H.~J. Jeon, M.~Li, K.~Srinivasan, A.~Mandlekar, A.~Garg, J.~Bohg, and D.~Sadigh.
\newblock Learning latent actions to control assistive robots.
\newblock {\em Autonomous robots}, 46(1):115--147, 2022.

\bibitem{lowe2017multi}
R.~Lowe, A.~Tamar, J.~Harb, O.~Pieter~Abbeel, and I.~Mordatch.
\newblock Multi-agent actor-critic for mixed cooperative-competitive environments.
\newblock {\em Advances in neural information processing systems}, 30, 2017.

\bibitem{maggio2024clio}
D.~Maggio, Y.~Chang, N.~Hughes, M.~Trang, D.~Griffith, C.~Dougherty, E.~Cristofalo, L.~Schmid, and L.~Carlone.
\newblock Clio: Real-time task-driven open-set 3d scene graphs.
\newblock {\em arXiv preprint arXiv:2404.13696}, 2024.

\bibitem{mastrogiuseppe2019spatiotemporal}
M.~Mastrogiuseppe, N.~Bertelsen, M.~F. Bedeschi, and S.~A. Lee.
\newblock The spatiotemporal organization of episodic memory and its disruption in a neurodevelopmental disorder.
\newblock {\em Scientific reports}, 9(1):18447, 2019.

\bibitem{min2022film}
S.~Y. Min, D.~S. Chaplot, P.~K. Ravikumar, Y.~Bisk, and R.~Salakhutdinov.
\newblock Film: Following instructions in language with modular methods.
\newblock In {\em International Conference on Learning Representations}, 2022.

\bibitem{natarajan2020effects}
M.~Natarajan and M.~Gombolay.
\newblock Effects of anthropomorphism and accountability on trust in human robot interaction.
\newblock In {\em Proceedings of the 2020 ACM/IEEE international conference on human-robot interaction}, pages 33--42, 2020.

\bibitem{nikolaidis2015efficient}
S.~Nikolaidis, R.~Ramakrishnan, K.~Gu, and J.~Shah.
\newblock Efficient model learning from joint-action demonstrations for human-robot collaborative tasks.
\newblock In {\em Proceedings of the tenth annual ACM/IEEE international conference on human-robot interaction}, pages 189--196, 2015.

\bibitem{niu2021multi}
Y.~Niu, R.~R. Paleja, and M.~C. Gombolay.
\newblock Multi-agent graph-attention communication and teaming.
\newblock In {\em AAMAS}, volume~21, page 20th, 2021.

\bibitem{nuxoll2007extending}
A.~M. Nuxoll and J.~E. Laird.
\newblock Extending cognitive architecture with episodic memory.
\newblock In {\em AAAI}, pages 1560--1564, 2007.

\bibitem{openai2023gpt4}
OpenAI.
\newblock Gpt-4 technical report, 2023.

\bibitem{packer2023memgpt}
C.~Packer, V.~Fang, S.~Patil, K.~Lin, S.~Wooders, and J.~Gonzalez.
\newblock Memgpt: Towards llms as operating systems.
\newblock {\em arXiv preprint}, 2023.

\bibitem{park2023generative}
J.~S. Park, J.~C. O'Brien, C.~J. Cai, M.~R. Morris, P.~Liang, and M.~S. Bernstein.
\newblock Generative agents: Interactive simulacra of human behavior.
\newblock {\em arXiv preprint arXiv:2304.03442}, 2023.

\bibitem{SMPL-X:2019}
G.~Pavlakos, V.~Choutas, N.~Ghorbani, T.~Bolkart, A.~A.~A. Osman, D.~Tzionas, and M.~J. Black.
\newblock Expressive body capture: {3D} hands, face, and body from a single image.
\newblock In {\em Proceedings IEEE Conf. on Computer Vision and Pattern Recognition (CVPR)}, pages 10975--10985, 2019.

\bibitem{puigwatch}
X.~Puig, T.~Shu, S.~Li, Z.~Wang, Y.-H. Liao, J.~B. Tenenbaum, S.~Fidler, and A.~Torralba.
\newblock Watch-and-help: A challenge for social perception and human-ai collaboration.
\newblock In {\em International Conference on Learning Representations}, 2021.

\bibitem{puig2024habitat}
X.~Puig, E.~Undersander, A.~Szot, M.~D. Cote, T.-Y. Yang, R.~Partsey, R.~Desai, A.~Clegg, M.~Hlavac, S.~Y. Min, et~al.
\newblock Habitat 3.0: A co-habitat for humans, avatars, and robots.
\newblock In {\em The Twelfth International Conference on Learning Representations}, 2024.

\bibitem{puig2023habitat}
X.~Puig, E.~Undersander, A.~Szot, M.~D. Cote, T.-Y. Yang, R.~Partsey, R.~Desai, A.~W. Clegg, M.~Hlavac, S.~Y. Min, et~al.
\newblock Habitat 3.0: A co-habitat for humans, avatars and robots.
\newblock {\em arXiv preprint arXiv:2310.13724}, 2023.

\bibitem{radford2021learning}
A.~Radford, J.~W. Kim, C.~Hallacy, A.~Ramesh, G.~Goh, S.~Agarwal, G.~Sastry, A.~Askell, P.~Mishkin, J.~Clark, et~al.
\newblock Learning transferable visual models from natural language supervision.
\newblock In {\em International conference on machine learning}, pages 8748--8763. PMLR, 2021.

\bibitem{ramakrishnan2022poni}
S.~K. Ramakrishnan, D.~S. Chaplot, Z.~Al-Halah, J.~Malik, and K.~Grauman.
\newblock Poni: Potential functions for objectgoal navigation with interaction-free learning.
\newblock In {\em Proceedings of the IEEE/CVF Conference on Computer Vision and Pattern Recognition}, pages 18890--18900, 2022.

\bibitem{rana2023sayplan}
K.~Rana, J.~Haviland, S.~Garg, J.~Abou-Chakra, I.~D. Reid, and N.~Suenderhauf.
\newblock Sayplan: Grounding large language models using 3d scene graphs for scalable task planning.
\newblock {\em CoRR}, 2023.

\bibitem{ravi2024sam2segmentimages}
N.~Ravi, V.~Gabeur, Y.-T. Hu, R.~Hu, C.~Ryali, T.~Ma, H.~Khedr, R.~Rädle, C.~Rolland, L.~Gustafson, E.~Mintun, J.~Pan, K.~V. Alwala, N.~Carion, C.-Y. Wu, R.~Girshick, P.~Dollár, and C.~Feichtenhofer.
\newblock Sam 2: Segment anything in images and videos, 2024.

\bibitem{rozo2016learning}
L.~Rozo, S.~Calinon, D.~G. Caldwell, P.~Jimenez, and C.~Torras.
\newblock Learning physical collaborative robot behaviors from human demonstrations.
\newblock {\em IEEE Transactions on Robotics}, 32(3):513--527, 2016.

\bibitem{samvelyan2019starcraft}
M.~Samvelyan, T.~Rashid, C.~Schroeder~de Witt, G.~Farquhar, N.~Nardelli, T.~G. Rudner, C.-M. Hung, P.~H. Torr, J.~Foerster, and S.~Whiteson.
\newblock The starcraft multi-agent challenge.
\newblock In {\em Proceedings of the 18th International Conference on Autonomous Agents and MultiAgent Systems}, pages 2186--2188, 2019.

\bibitem{schwartz2012overview}
S.~H. Schwartz.
\newblock An overview of the schwartz theory of basic values.
\newblock {\em Online readings in Psychology and Culture}, 2(1):11, 2012.

\bibitem{shafiullah2022clip}
N.~M.~M. Shafiullah, C.~Paxton, L.~Pinto, S.~Chintala, and A.~Szlam.
\newblock Clip-fields: Weakly supervised semantic fields for robotic memory.
\newblock {\em arXiv preprint arXiv:2210.05663}, 2022.

\bibitem{sharon2015conflict}
G.~Sharon, R.~Stern, A.~Felner, and N.~R. Sturtevant.
\newblock Conflict-based search for optimal multi-agent pathfinding.
\newblock {\em Artificial intelligence}, 219:40--66, 2015.

\bibitem{shi2024replug}
W.~Shi, S.~Min, M.~Yasunaga, M.~Seo, R.~James, M.~Lewis, L.~Zettlemoyer, and W.-t. Yih.
\newblock Replug: Retrieval-augmented black-box language models.
\newblock In {\em Proceedings of the 2024 Conference of the North American Chapter of the Association for Computational Linguistics: Human Language Technologies (Volume 1: Long Papers)}, pages 8364--8377, 2024.

\bibitem{shinn2024reflexion}
N.~Shinn, F.~Cassano, A.~Gopinath, K.~Narasimhan, and S.~Yao.
\newblock Reflexion: Language agents with verbal reinforcement learning.
\newblock {\em Advances in Neural Information Processing Systems}, 36, 2024.

\bibitem{suarez2019neural}
J.~Suarez, Y.~Du, P.~Isola, and I.~Mordatch.
\newblock Neural mmo: A massively multiagent game environment for training and evaluating intelligent agents.
\newblock {\em arXiv preprint arXiv:1903.00784}, 2019.

\bibitem{sumers2023cognitive}
T.~Sumers, S.~Yao, K.~Narasimhan, and T.~L. Griffiths.
\newblock Cognitive architectures for language agents.
\newblock {\em arXiv preprint arXiv:2309.02427}, 2023.

\bibitem{sun2023think}
J.~Sun, C.~Xu, L.~Tang, S.~Wang, C.~Lin, Y.~Gong, H.-Y. Shum, and J.~Guo.
\newblock Think-on-graph: Deep and responsible reasoning of large language model with knowledge graph.
\newblock {\em arXiv preprint arXiv:2307.07697}, 2023.

\bibitem{szot2023adaptive}
A.~Szot, U.~Jain, D.~Batra, Z.~Kira, R.~Desai, and A.~Rai.
\newblock Adaptive coordination in social embodied rearrangement.
\newblock In {\em International Conference on Machine Learning}, pages 33365--33380. PMLR, 2023.

\bibitem{tenenbaum2011grow}
J.~B. Tenenbaum, C.~Kemp, T.~L. Griffiths, and N.~D. Goodman.
\newblock How to grow a mind: Statistics, structure, and abstraction.
\newblock {\em science}, 331(6022):1279--1285, 2011.

\bibitem{tsoi2020sean}
N.~Tsoi, M.~Hussein, J.~Espinoza, X.~Ruiz, and M.~V{\'a}zquez.
\newblock Sean: Social environment for autonomous navigation.
\newblock In {\em Proceedings of the 8th international conference on human-agent interaction}, pages 281--283, 2020.

\bibitem{tulving1972episodic}
E.~Tulving.
\newblock Episodic and semantic memory.
\newblock {\em Organization of memory/Academic Press}, 1972.

\bibitem{tulving1983elements}
E.~Tulving.
\newblock Elements of episodic memory, 1983.

\bibitem{wang2023voyager}
G.~Wang, Y.~Xie, Y.~Jiang, A.~Mandlekar, C.~Xiao, Y.~Zhu, L.~Fan, and A.~Anandkumar.
\newblock Voyager: An open-ended embodied agent with large language models.
\newblock {\em arXiv preprint arXiv:2305.16291}, 2023.

\bibitem{wang2023describe}
Z.~Wang, S.~Cai, G.~Chen, A.~Liu, X.~Ma, and Y.~Liang.
\newblock Describe, explain, plan and select: interactive planning with llms enables open-world multi-task agents.
\newblock In {\em Thirty-seventh Conference on Neural Information Processing Systems}, 2023.

\bibitem{wang2024karma}
Z.~Wang, B.~Yu, J.~Zhao, W.~Sun, S.~Hou, S.~Liang, X.~Hu, Y.~Han, and Y.~Gan.
\newblock Karma: Augmenting embodied ai agents with long-and-short term memory systems.
\newblock {\em arXiv preprint arXiv:2409.14908}, 2024.

\bibitem{wen2022multi}
M.~Wen, J.~Kuba, R.~Lin, W.~Zhang, Y.~Wen, J.~Wang, and Y.~Yang.
\newblock Multi-agent reinforcement learning is a sequence modeling problem.
\newblock {\em Advances in Neural Information Processing Systems}, 35:16509--16521, 2022.

\bibitem{weston2014memory}
J.~Weston, S.~Chopra, and A.~Bordes.
\newblock Memory networks.
\newblock {\em arXiv preprint arXiv:1410.3916}, 2014.

\bibitem{yang2024snapmem}
Y.~Yang, H.~Yang, J.~Zhou, P.~Chen, H.~Zhang, Y.~Du, and C.~Gan.
\newblock Snapmem: Snapshot-based 3d scene memory for embodied exploration and reasoning.
\newblock {\em arXiv preprint arXiv:2411.17735}, 2024.

\bibitem{yasunaga2023retrieval}
M.~Yasunaga, A.~Aghajanyan, W.~Shi, R.~James, J.~Leskovec, P.~Liang, M.~Lewis, L.~Zettlemoyer, and W.-T. Yih.
\newblock Retrieval-augmented multimodal language modeling.
\newblock In {\em International Conference on Machine Learning}, pages 39755--39769. PMLR, 2023.

\bibitem{yu2024mineland}
X.~Yu, J.~Fu, R.~Deng, and W.~Han.
\newblock Mineland: Simulating large-scale multi-agent interactions with limited multimodal senses and physical needs.
\newblock {\em arXiv preprint arXiv:2403.19267}, 2024.

\bibitem{zhang2023building}
H.~Zhang, W.~Du, J.~Shan, Q.~Zhou, Y.~Du, J.~B. Tenenbaum, T.~Shu, and C.~Gan.
\newblock Building cooperative embodied agents modularly with large language models, 2023.

\bibitem{zhang2018semantic}
L.~Zhang, L.~Wei, P.~Shen, W.~Wei, G.~Zhu, and J.~Song.
\newblock Semantic slam based on object detection and improved octomap.
\newblock {\em IEEE Access}, 6:75545--75559, 2018.

\bibitem{zhang2025surveygraphretrievalaugmentedgeneration}
Q.~Zhang, S.~Chen, Y.~Bei, Z.~Yuan, H.~Zhou, Z.~Hong, J.~Dong, H.~Chen, Y.~Chang, and X.~Huang.
\newblock A survey of graph retrieval-augmented generation for customized large language models, 2025.

\bibitem{zheng2023asystem}
K.~Zheng, A.~Paul, and S.~Tellex.
\newblock Asystem for generalized 3d multi-object search.
\newblock In {\em 2023 IEEE International Conference on Robotics and Automation (ICRA)}, pages 1638--1644. IEEE, 2023.

\bibitem{zhou2023long}
F.~Zhou, H.~Liu, H.~Zhao, and L.~Liang.
\newblock Long-term object search using incremental scene graph updating.
\newblock {\em Robotica}, 41(3):962--975, 2023.

\bibitem{vico}
Q.~Zhou, H.~Zhang, X.~Lin, Z.~Zhang, Y.~Chen, W.~Liu, Z.~Zhang, S.~Chen, L.~Fang, Q.~Lyu, X.~Sun, J.~Yang, Z.~Wang, B.~C. Dang, Z.~Chen, D.~Ladia, J.~Liu, and C.~Gan.
\newblock Virtual community: An open world for humans, robots, and society.
\newblock 2025.

\bibitem{zhou2024sotopia}
X.~Zhou, H.~Zhu, L.~Mathur, R.~Zhang, H.~Yu, Z.~Qi, L.-P. Morency, Y.~Bisk, D.~Fried, G.~Neubig, et~al.
\newblock Sotopia: Interactive evaluation for social intelligence in language agents.
\newblock In {\em The Twelfth International Conference on Learning Representations}, 2024.

\end{thebibliography}
